\documentclass[sn-vancouver, Numbered]{sn-jnl}

\usepackage{float}
\usepackage{longtable}
\usepackage{adjustbox}
\usepackage{pdflscape}
\usepackage{geometry}

\usepackage{graphicx}%
\usepackage{multirow}%
\usepackage{amsmath,amssymb,amsfonts}%
\usepackage{amsthm}%
\usepackage{mathrsfs}%
\usepackage[title]{appendix}%
\usepackage{textcomp}%
\usepackage{manyfoot}%
\usepackage{booktabs}%
\usepackage{algorithm}%
\usepackage{algorithmicx}%
\usepackage{algpseudocode}%
\usepackage{listings}%
\usepackage[table,xcdraw]{xcolor}

\theoremstyle{thmstyleone}%
\theoremstyle{thmstyletwo}%

\theoremstyle{thmstylethree}%

\begin{document}

\title[Article Title]{An Unsupervised Natural Language Processing Pipeline for Assessing Referral Appropriateness}


\author*[1]{\fnm{Vittorio} \sur{Torri} \dgr{PhD}}\email{vittorio.torri@polimi.it}

\author[2]{\fnm{Annamaria} \sur{Bottelli} \dgr{MSc}}

\author[2]{\fnm{Michele} \sur{Ercolanoni}~\dgr{MSc}}

\author[3]{\fnm{Olivia} \sur{Leoni} \dgr{PhD}}

\author[1,4]{\fnm{Francesca} \sur{Ieva}~\dgr{PhD}}

\affil*[1]{\orgdiv{MOX - Modelling and Scientific Computing Lab, Department of Mathematics},
\orgname{Politecnico di Milano}, \orgaddress{\street{Piazza Leonardo da Vinci 32}, \city{Milan}, \postcode{20133}, \country{Italy}}}

\affil[2]{\orgname{ARIA s.p.a - Azienda Regionale per l’Innovazione e gli Acquisti}, \orgaddress{\street{Via Taramelli 26}, \city{Milan}, \postcode{20124},  \country{Italy}}}

\affil[3]{\orgdiv{U.O. Osservatorio Epidemiologico, DG Welfare}, \orgname{Regione Lombardia}, \orgaddress{\street{Piazza Città di
Lombardia 1}, \city{Milan}, \postcode{20124}, \country{Italy}}}

\affil[4]{\orgdiv{HDS - Health Data Science Centre}, \orgname{Human Technopole}, \orgaddress{\street{Viale Rita Levi-Montalcini 1}, \city{Milan}, \postcode{20157}, \country{Italy}}}


 \abstract{\textbf{Background:} Assessing the appropriateness of diagnostic referrals is critical for improving healthcare efficiency and reducing unnecessary procedures. However, this task becomes challenging when referral reasons are recorded only as free text rather than structured codes. In the Italian National Health Service (NHS), referral motivations are systematically entered as unstructured text, making large-scale evaluation against official guidelines difficult to automate and scale. To address this gap, we propose a fully unsupervised Natural Language Processing (NLP) pipeline capable of extracting and evaluating referral reasons without relying on labelled datasets, which are costly and time-consuming to produce.\\
\textbf{Methods:} Our pipeline leverages Transformer-based embeddings pre-trained on Italian medical texts to semantically cluster referral reasons and assess their alignment with appropriateness guidelines. It operates in an unsupervised setting and is designed to generalize across different examination types. We analyzed two complete regional datasets from the Lombardy Region (Italy), covering all referrals between 2019 and 2021 for venous echocolordoppler of the lower limbs (ECD; n=496,971) and flexible endoscope colonoscopy (FEC; n=407,949). The pipeline was developed on the ECD dataset and tested on the FEC dataset. For both, a random sample of 1,000 referrals was manually annotated to measure performance. Results were also stratified by year, region, and physician type.\\
\textbf{Results:} The pipeline achieved high performance in identifying referral reasons (Precision=92.43\% (ECD), 93.59\% (FEC); Recall=83.28\% (ECD), 92.70\% (FEC)) and appropriateness (Precision=93.58\% (ECD), 94.66\% (FEC); Recall=91.52\% (ECD), 93.96\% (FEC)). Inappropriate referrals were significantly more prevalent for ECD (34.1\%) than for FEC (5.3\%). The analysis revealed geographical, temporal, and physician type variations, including a drop in inappropriate referrals during the COVID-19 pandemic in 2020. Additionally, several referral motivations not covered by current guidelines were identified, highlighting areas for policy improvement. These findings contributed to updates to the regional guidelines in Lombardy.\\
\textbf{Conclusions:} This study presents a robust, scalable, unsupervised NLP pipeline for assessing referral appropriateness from unstructured texts in large, real-world datasets. It demonstrates how such data can be effectively leveraged, providing public health authorities with a deployable AI tool to monitor practices and support evidence-based policy.
}
\keywords{Natural language processing, Artificial Intelligence, Machine Learning, Referrals, Unsupervised learning, Examination appropriateness, Guideline adherence, Public healthcare system, Quality of healthcare}


\pacs[MSC Classification]{68T50}

\maketitle

\section{Background}
\label{sec:background}
In Italy, the National Health Service (NHS) is the primary provider of healthcare services, offering universal coverage, with administration decentralized at the regional level~\cite{signorelli2020universal}. Despite its strong performance in international rankings, the Italian NHS has faced growing challenges in providing timely access to healthcare services in recent years, with increasing numbers of citizens turning to private providers~\cite{maietti2023changes}. 

One significant contributor to these delays is the issue of referrals appropriateness, a concern shared by many advanced healthcare systems~\cite{brownlee2017evidence, kale2013trends}. Inappropriate referrals lead to resource waste, increased waiting times, and delays in accurate diagnoses, ultimately reducing the effectiveness of the healthcare system~\cite{hyttinen2016systematic}. Within the Italian NHS, referrals from general practitioners (GPs) and specialists are required to book diagnostic examinations, making it crucial to assess the reasons behind these referrals to ensure their appropriateness.

When issuing a referral, Italian physicians must complete a standardized form, including a free-text field known as the \textit{Clinical Question}, where they specify the reason for the prescription. The unstructured nature of this field, coupled with the absence of standardized alternatives, poses a major challenge for systematic analysis. Addressing this issue requires Natural Language Processing (NLP) techniques. In this study, we frame the task as a clustering problem: clinical questions are grouped using Transformer-based embeddings~\cite{vaswani2017attention} to capture semantic similarity, and clusters are subsequently mapped to guideline-derived categories of appropriateness.

Despite its importance, research on NLP-driven analysis of Italian referrals is limited. Venturelli et al.~\cite{venturelli2021using} used proprietary commercial software (\textit{Clinika VAP}~\cite{clinikavap}) to classify referrals as appropriate or inappropriate, but without explicitly identifying the reasons for prescriptions. To the best of our knowledge, the only other study applying NLP on Italian referrals' texts is \cite{torri2024nlp}, which focused on extracting and analyzing follow-up waiting times from clinical questions rather than assessing referrals appropriateness.

Most previous studies on appropriateness in the Italian NHS relied on structured data or manual text analysis, focusing on medication prescriptions rather than medical examinations~\cite{cattaneo2020drug, segala2020antibiotic, franchi2020use}. Research on the appropriateness of examinations is more limited~\cite{gion2022use, lencioni2023looking} and has not incorporated textual data. Other studies focused on the development of decision-support systems for guideline adherence, both in primary care~\cite{camerotto2017ermete, calcaterra2018clinical} and hospital settings~\cite{roncato2019farmaprice, ghibelli2013prevention}. While effective in specific scenarios~\cite{gupta2014effect, mariotti2022consensus}, such tools are not suited for retrospective analyses of their performance or for quantifying the prevalence of appropriate or inappropriate referrals.

Internationally, several studies have employed NLP and rule-based systems to assess referral appropriateness, often in disease-specific contexts. Haddad et al.~\cite{haddad2024introducing} analyzed Russian referrals' texts for appropriateness using labelled data; Sagheb et al.~\cite{sagheb2022artificial} designed a rule-based algorithm to verify guidelines adherence for asthma patients from EHR documents; similarly in~\cite{kerr2015measuring} a rule-based approach was applied to assess guideline adherence for gout patients. Dimitrov et al.~\cite{dimitrov2024nationwide} conducted a nationwide evaluation of breast cancer guideline adherence in EHRs using proprietary commercial software (\textit{Danny platform}~\cite{dannyplatform}), which incorporated rules and manual reviews. Villeana et al.~\cite{villena2021supporting} analyzed Chilean referrals' texts using labelled data to binary classify them for correct priority classes. 

However, all these approaches rely on labeled data or proprietary software, limiting their adaptability and scalability across different examination types.

To address these gaps, this study proposes the first fully unsupervised NLP pipeline for analyzing appropriateness in Italian referrals. The pipeline is applicable to any examination for which guidelines define appropriate and inappropriate referral reasons. We developed this approach on a first case study related to venous echocolordoppler of the lower limbs (ECD) and we present results also on a second case study related to flexible endoscope colonoscopy (FEC). Both datasets include all referrals issued during three years in the entire the Lombardy Region, the most populated Italian region, with approximately 10 million inhabitants - akin to a small state. 

These two examinations were selected because both are characterized by a high prevalence of inappropriate referrals~\cite{petruzziello2012appropriateness, hassan2007appropriateness, zanatta2002appropriateness, stegher2017appropriateness} and had already been targeted by Lombardy’s 2015 resolutions on angiology and gastroenterology~\cite{dgrAngio, dgrGastro}.
However, no comprehensive study has assessed the impact of those resolutions at scale. Our analysis provides, for the first time, a regional-scale evaluation of referral appropriateness and of guideline effectiveness, which informed the Lombardy Region’s December 2023 resolution~\cite{dgr2023}.

This work thus provides both a methodological contribution, by introducing a scalable NLP pipeline for appropriateness assessment, and an applied contribution, by generating evidence that informed regional policies.

The remainder of this paper is structured as follows: Section \ref{sec:methods} describes the datasets and the pipeline, Section \ref{sec:results} presents the results, Section \ref{sec:discussion} addresses implications, limitations, and future developments; and Section \ref{sec:conclusions} concludes.

\section{Methods}
\label{sec:methods}
\subsection{Data}
\subsubsection{Referrals datasets}
\begin{figure}
    \centering
    \includegraphics[width=\textwidth]{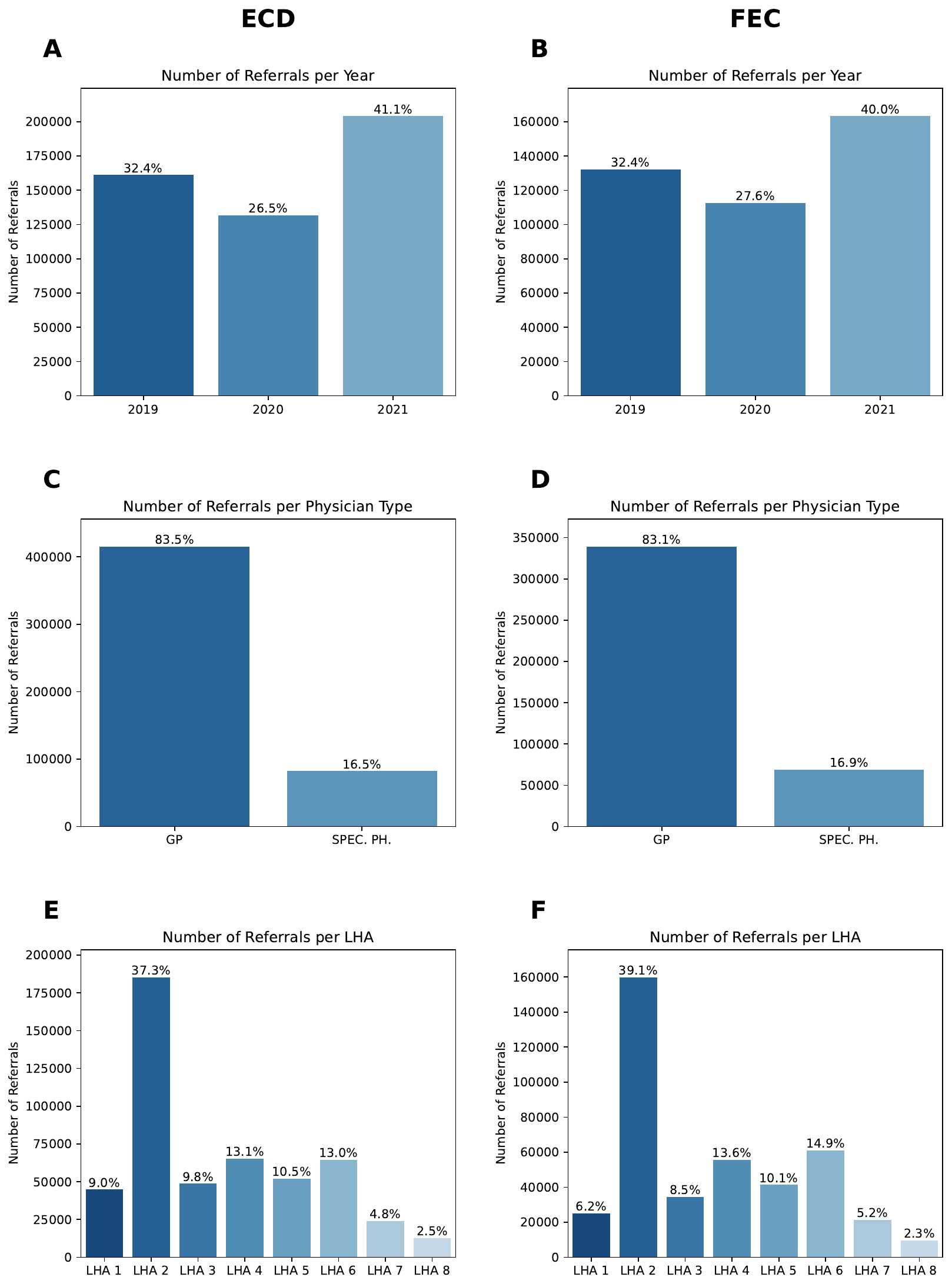}
    \caption{Referrals distributions across years (A, B), type of prescribing physicians (C, D) and LHA (E, F) for ECD (A,C,E) and FEC (B,D,F).}
    \label{fig:desc-barplots}
\end{figure}

The datasets analyzed in this study are extracted from all referrals issued in the Lombardy Region between 2019 and 2021 for ECD and FEC, totalling 496,971 referrals for ECD and 407,949 for FEC. Each referral includes the clinical question, the year, the Local Health Authority (LHA) of the prescribing physician and the type of prescribing physician. Physician types are categorized as general practitioners (GPs) or specialized physicians, while LHAs are the eight healthcare administrative areas within the Lombardy Region. Figure~\ref{fig:desc-barplots} illustrates the distribution of referrals among these groups.

The analysis focuses on the clinical question (CQ), a free-text field typically containing one or a few sentences. These texts exhibit frequent spelling and typing errors, heavy use of abbreviations, inconsistent punctuation, and non-standard capitalization. The dataset for ECD includes 229,509 distinct CQ, with only 15\% appearing more than once, while the dataset for FEC includes 89,734 distinct CQ, with 41\% appearing more than once. The median length for ECD is 71 characters (IQR=39), while for FEC it is 57 (IQR=20). Only 189 for ECD~(0.08\%) and 137 for FEC~(0.15\%) contain a meaningful ICD code, underscoring the necessity of employing NLP for this analysis.

The clinical questions do not contain personal information about patients or physicians. 

\subsubsection{Manual annotations}
To validate the pipeline, a randomly selected subset of 1,000 distinct clinical questions was manually labelled for both ECD and FEC. The labels correspond to categories representing diseases or symptoms that prompted the referral. During an initial review of this subset, ambiguous cases were identified, and an angiologist and a gastroenterologist were consulted to resolve these doubts and align the clusters with prescription guidelines. Annotation guidelines were subsequently refined, and the entire subset was annotated accordingly. Two other reviews of the annotations corrected remaining errors and clarified uncertainties, ensuring the final labelled datasets were error-free. More details on the manual annotations are provided in Additional File 1, Section A.

\subsubsection{Prescription guidelines}
\begin{figure}
    \centering
    \includegraphics[width=\linewidth]{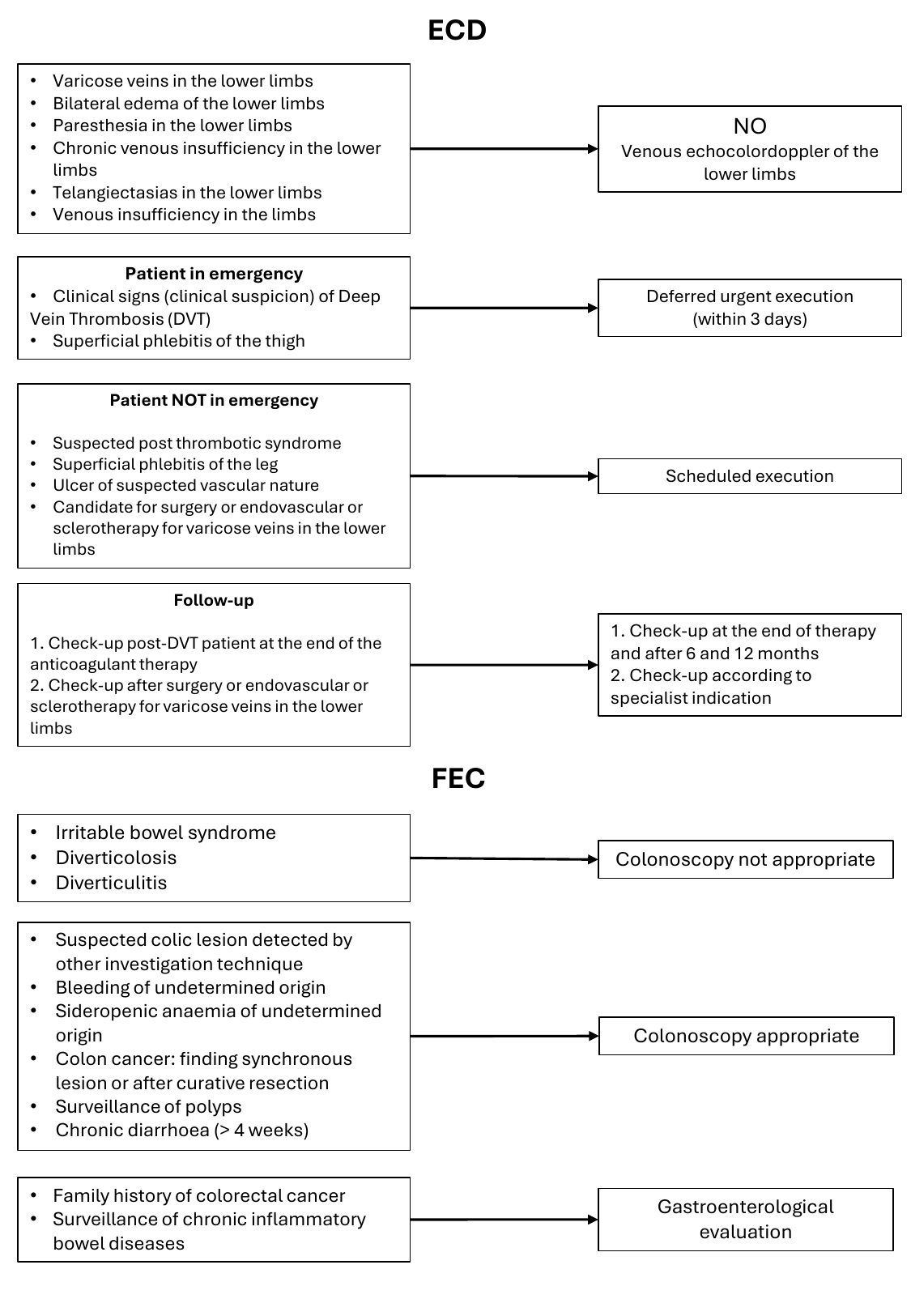}
    \caption{Guidelines for referrals for ECD and FEC in the Lombardy Region. Derived from the the Appendix 3 of \cite{dgrAngio} and the Appendix 2 of \cite{dgrGastro}, translated into English.}
    \label{fig:guidelines}
\end{figure}

The guidelines used as a reference to assess referral appropriateness are extracted from~\cite{dgrAngio} for ECD and from~\cite{dgrGastro} for FEC. In particular, we extracted the appropriate and inappropriate diseases and symptoms, considering separately also cases that are more likely to be appropriate or more likely to be inappropriate, depending on details that might not be available in clinical questions. English schematized summaries of these guidelines are reported in Figure~\ref{fig:guidelines}.

\subsection{Evaluation of prescription appropriateness}
In this section, we present our pipeline to evaluate prescription appropriateness. The process can be divided into two main steps:
\begin{enumerate}
    \item Identify the reason for the referral: extract the central semantic meaning from the clinical question.
    \item Verify appropriateness: determine if the identified reason aligns with established guidelines.
\end{enumerate}

The absence of labels for this type of data prevented us from adopting a supervised approach. A rule-based approach was excluded since we aim to build a pipeline that can generalize to any examination, and it would not be feasible to develop tailored rules for each type of examination. Moreover, we are interested in understanding the reasons for these referrals, which might include additional and/or more detailed categories than those mentioned in the guidelines. Because of this, the problem was addressed as a clustering problem.

\begin{figure}[h]
    \centering
    \includegraphics[width=\textwidth]{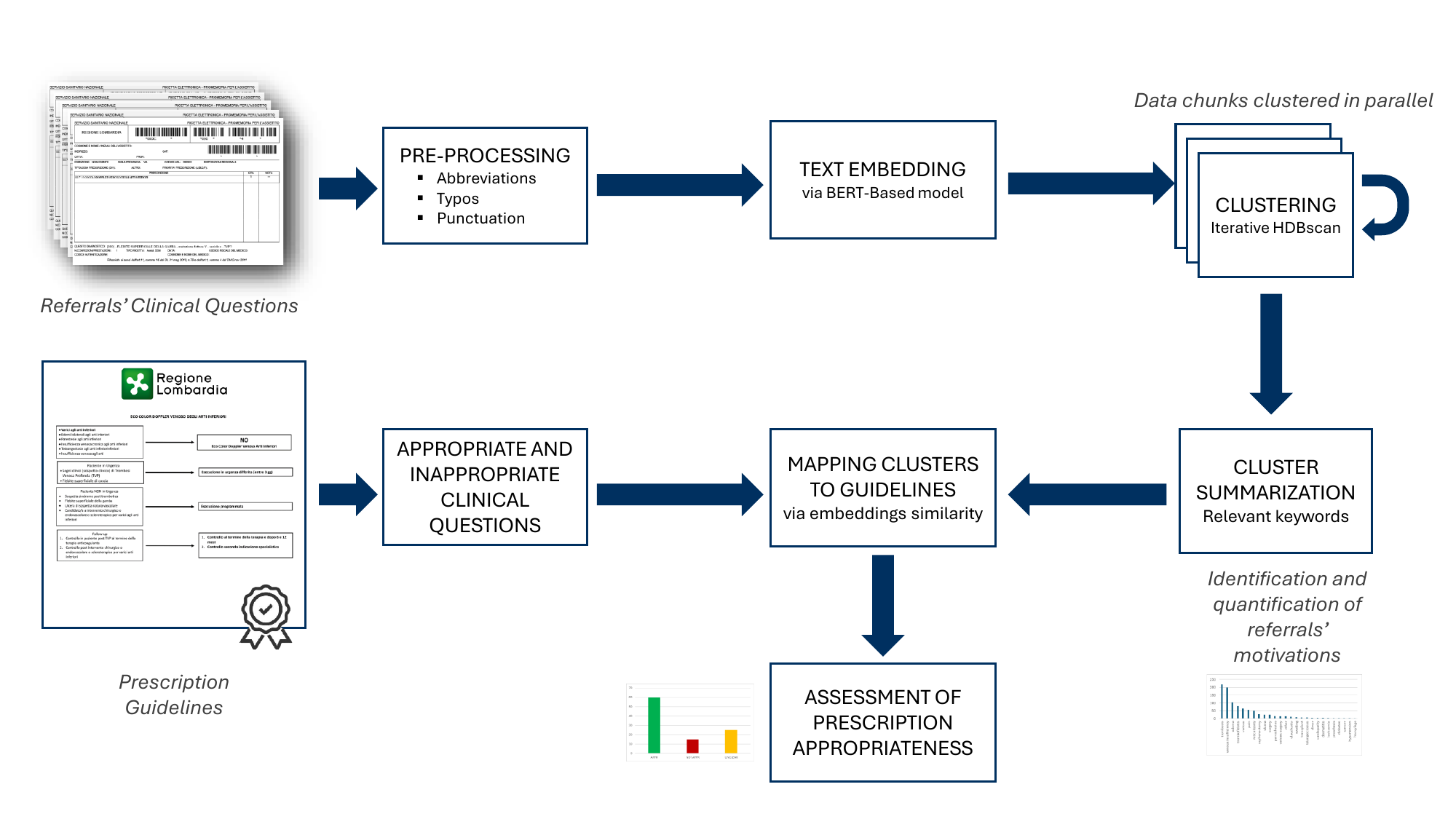}
    \caption{Schema of the pipeline for assessing referrals reasons and appropriateness}
    \label{fig:pipeline}
\end{figure}

Figure \ref{fig:pipeline} depicts our pipeline with the blocks that compose it, and below, we detail each block.

\subsubsection{Pre-processing}
Physicians write referrals during patients' examinations, with limited time available. This leads them to write clinical questions that are far from well-readable and correct texts. The grammar of the sentences is often poor, there are frequent typing errors, and the punctuation is either not used or used incorrectly. The use of uppercase letters does not have any particular meaning, and the use of abbreviations is very frequent. Even in the two types of examination analyzed in this study, multiple abbreviations can indicate the same disease/symptoms. In the manually labelled subsets, 15\% of clinical questions were marked as having typing errors and 68\% as containing abbreviations.

To address these issues, suitable pre-processing steps are applied:
\begin{enumerate}
    \item lowercasing
    \item punctuation removal
    \item abbreviation expansion
    \item (partial) typos correction
\end{enumerate}
We remove all punctuation and lowercase sentences due to the inconsistent use of these elements. 

For abbreviation expansion, we extract all the unique words with length~$\leq~3$, and we identify those corresponding to abbreviations of medical terms, defining a vocabulary for their replacement in the texts. 

To fix the typing mistakes, all words appearing with frequency~$<~1\%$ and having another word similar to them with frequency~$>~1\%$ are selected. These words are considered misspelt, and the most similar word with frequency~$>~1\%$ is chosen as their correction. The similarity is measured with edit distance, and the minimum threshold is chosen by minimizing errors in a manually reviewed sample of 100 corrections.

Examples of pre-processing effects and an ablation analysis are reported in Additional File 1, Section B.

\subsubsection{Text embedding}
\label{sec:bert-pretraining}
To appropriately cluster the clinical questions, it is necessary to produce a representation capturing the central semantics of these texts without getting confused by their widely present noise. 

Considering the many synonyms and composite terms of the medical jargon used in these texts, vocabulary-based representations, such as TF-IDF~\cite{leskovec2020mining}, cannot be expected to work well for this task.

Static word embeddings, such as Word2Vec~\cite{mikolov2013efficient}, can capture the semantic similarities between words with similar meanings (e.g., vasculopathy - arteriopathy, phlebitis - thrombophlebitis) differently from vocabulary-based representations. Nevertheless, they might have more difficulty distinguishing between relevant and irrelevant tokens in our sentences since they do not consider the context in which tokens appear. 

On the contrary, context-based embeddings produced by transformers model consider the surrounding context while embedding each token and have often shown superior performances in the recent clinical NLP literature~\cite{si2019enhancing, beaney2024comparing, zhang2019extracting, saha2020understanding}. The problem is particularly challenging in our case due to the limited number of embedding models available for the Italian language, especially for the clinical domain.

To this purpose, we adopt the same model developed in~\cite{torri2024weakly} to cluster sentences extracted from Italian discharge summaries, and we verify that even with referrals' clinical questions, its performances are superior to other models. This model consists of a fine-tuned version of the last Italian version of BERT, Umberto~\cite{tamburini2020bertology}, on a dataset of publicly available medical documents in Italian and other languages (translated into Italian)~\cite{magnini2022european}. 

\subsubsection{Clustering} 
Some studies have shown that HDBSCAN often outperforms other clustering algorithms—such as K-Means, hierarchical clustering, and DBSCAN-when applied to text embeddings, particularly when preceded by a dimensionality reduction step~\cite{asyaky2021improving, saha2023influence}. 
Text embeddings typically lie on complex, non-linear manifolds and may form clusters with varying densities and irregular shapes.
Algorithms like K-Means assume convex cluster geometries and require predefining the number of clusters, which is impractical for our task. Hierarchical clustering is sensitive to noise and outliers and lacks the ability to leave out unclustered data points, often producing forced or less meaningful groupings. In contrast, HDBSCAN does not require the number of clusters to be specified in advance and allows some data points to remain unclustered, making it a better fit for our scenario. Based on these advantages and on a comparison of the results on the ECD dataset, we selected HDBSCAN for the clustering step. 

HDBSCAN requires specifying some hyperparameters, including the minimum cluster size, the minimum number of samples in the neighborhood of a core point ($min\_samples$), and the epsilon. To better accommodate the varying sizes of clusters  in our dataset, we adopted a recursive strategy: the minimum cluster size was iteratively reduced, halving at each step, and applied only to data points that remained unclustered in the previous iteration. For the dimensionality reduction, we compared PCA, UMAP and no dimensionality reduction.

We performed a grid search and we selected the configuration with PCA with 80\% of variance threshold, reducing the minimum cluster size from 500 to 10, with $min\_samples=5$ and $\epsilon=0.1$. The complete grid search results, 
together with comparisons to k-means and agglomerative hierarchical clustering, are reported in Additional File~1, Section~C.

Due to resource constraints, it was not feasible to run HDBSCAN on the entire datasets. Therefore, we divided the data into chunks of 30,000 clinical questions and clustered them independently. In later stages, the summarization and mapping procedures allow to merge clusters from different chunks that correspond to the same diseases or symptoms.

\subsubsection{Cluster summarization}
Summarizing the identified clusters is necessary to understand the symptoms or diseases corresponding to each cluster. To automatize this step, we extracted keywords from each cluster. The most frequent words in each cluster were initially selected, considering a decreasing frequency threshold from 80\% until at least two keywords were found. Among these, we excluded those with high frequency in at least half of the clusters. This allowed us to exclude not only stop words but also frequent words for the specific examination we are analyzing.

\subsubsection{Mapping clusters to guidelines}
After summarization, the keyword representation of the clusters is used to map them to the guidelines. The process involves:
\begin{enumerate}
    \item Computing embeddings for each clinical question described in the guidelines.
    \item Generating an embedding for each cluster based on its keyword summary, enabling direct comparison with the guideline embeddings.
    \item Comparing cluster embeddings with guideline embeddings using cosine similarity. Clusters were assigned to the guideline entry with the highest similarity, provided there is one with similarity $> 0.50$.
\end{enumerate}

\subsection{Experiments and Evaluation}
\begin{figure}
    \centering
    \includegraphics[width=\linewidth]{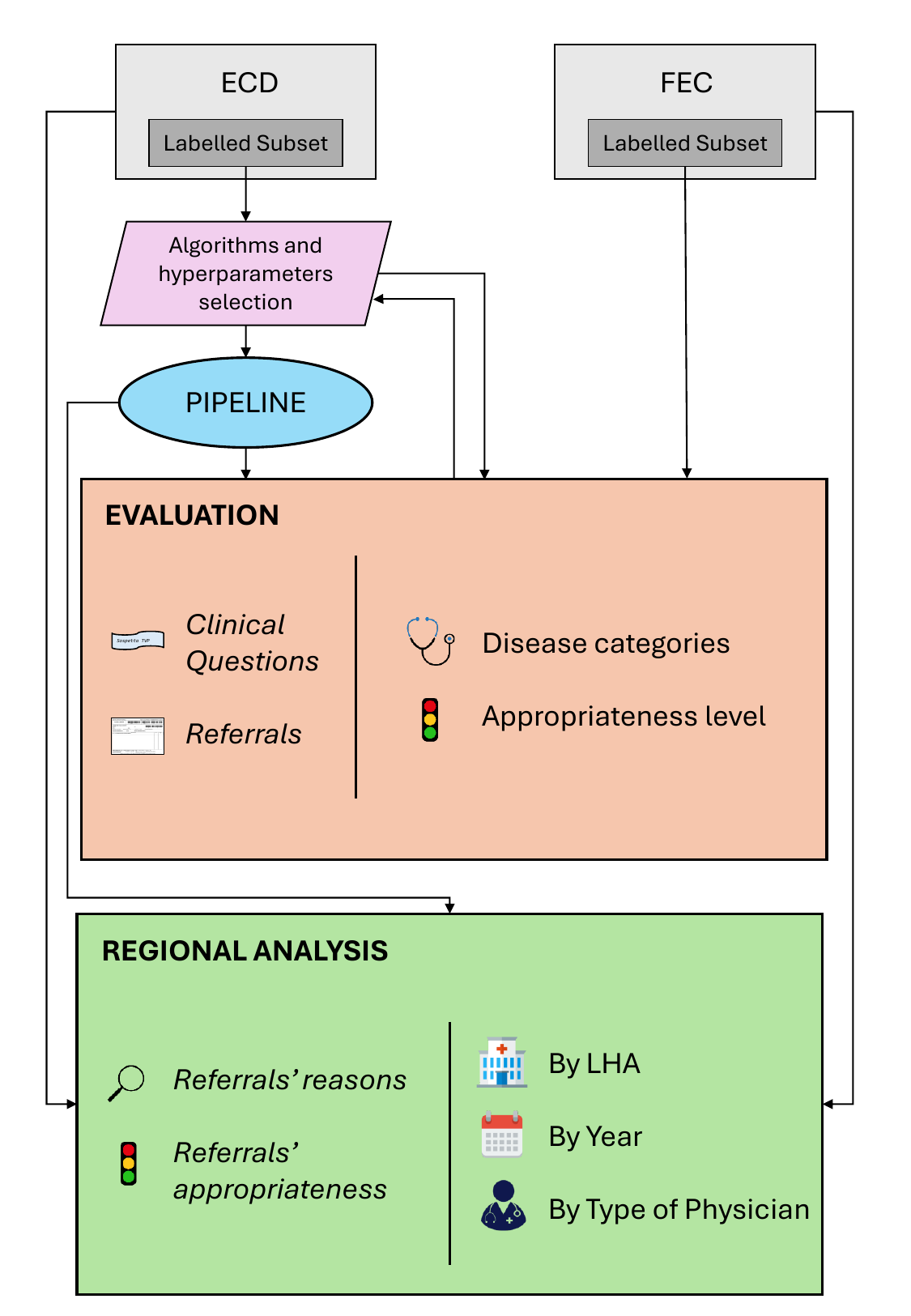}
    \caption{Summary of the experiments and evaluation procedure}
    \label{fig:eval_proc}
\end{figure}

In this section, we detail the experimental settings and the evaluation procedure for our pipeline, which are summarized in Figure~\ref{fig:eval_proc}.

\subsubsection{Experimental Setup and Constraints}
All experiments were conducted in an offline environment with limited computational resources:  16 GB of RAM, 16 cores Intel Xeon CPU, no GPU. For security and privacy reasons the data could not be move out of this environment. These constraints, commonly present in healthcare settings, influenced several pipeline design decisions, such as splitting the dataset into chunks for clustering, avoiding re-training Tranformer-based models and avoiding the use of generative large language models.

\subsubsection{Evaluation Procedure}
The pipeline was evaluated by comparing its output with the labels on the manually annotated subsets. Rather than applying the pipeline solely to the labelled records—which would hinder clustering performance due to limited data—we ran the full pipeline on the entire datasets and then extracted results for the annotated subsets. This ensured meaningful cluster formation and fair evaluation.

We performed evaluation at two levels:

\begin{itemize}
    \item \textbf{Disease label evaluation}: We assessed the accuracy of the disease labels produced by the pipeline by comparing them to ground-truth labels. To reduce noise, we retained only labels with more than 10 occurrences (1\%) in the labelled subset, grouping all others into an \textit{Other} category. Clusters not mapped to any label or unclustered clinical questions were assigned to the \textit{Other} category. Metrics were computed at:
    \begin{itemize}
        \item \textit{Aggregate level}: to measure overall precision, recall, and F1-score across all labels.
        \item \textit{Per-label level}: to evaluate performance on individual categories.
    \end{itemize}
    
    \item \textbf{Appropriateness label evaluation}: The quality of the pipeline’s assignment of appropriateness labels was evaluated using a four-class scheme: \textit{appropriate}, \textit{inappropriate}, \textit{likely (in)appropriate}, and \textit{undetermined}. The \textit{likely (in)appropriate} category included referrals where reasons mentioned in the guidelines were present, but the text lacked complete information to assess the (in)appropriateness. The \textit{undetermined} category was used for referrals not addressed in the clinical guidelines.
    Also in this case, we computed metrics and the \textit{aggregate} and \textit{per-label} levels.
\end{itemize}

At both levels, we computed the metrics at the single clinical questions level and weighting clinical questions by their corresponding referral counts. The latter allows to incorporating the datasets' real-world distributions into the analysis.

We also qualitatively show t-SNE 2D projections of the embeddings of the labelled clinical questions. Although this type of visualization has known limitations~\cite{liu2017visual}, it remains a common approach in the literature for obtaining a qualitative overview of embedding structures~\cite{oubenali2022visualization}.

\subsubsection{Hyperparameter Selection and Validation Strategy}
Clustering algorithm selection and hyperparameter tuning was carried out exclusively on the ECD dataset. The FEC dataset was reserved for external validation only, ensuring that all performance results reported on FEC represent generalization to unseen data and are free from overfitting to the development set.

\subsubsection{Comparative Evaluation}
To benchmark our pipeline, we compared it with alternative approaches using different text representations and/or clustering methods. These included:
\begin{itemize}
    \item Embeddings from the original UmBERTo model (baseline)
    \item Embeddings from an Italian Word2Vec model~\cite{di2021analysis}
    \item TF-IDF vector representations
    \item Latent Dirichlet Allocation (LDA)
    \item Direct matching between embeddings of clinical questions and embeddings of diseases from the guidelines
\end{itemize}
Additional details are provided in Additional File 1, Section C.

\subsubsection{Regional and Temporal Analysis}
Finally, we applied the pipeline to the full datasets and reported results at a regional level. This provides a complete quantification of referrals for each identified category, together with their appropriateness. Moreover, we present an enrichment analysis stratifying results by year (pre-, during, and post-COVID pandemic), Local Health Authority (LHA), and prescribing physician type (general practitioners vs. specialists).
\section{Results}
\label{sec:results}
\begin{table}[tbh]
\centering
\caption{Summary evaluation on the disease labels}
\label{tab:clusters-res}
\begin{tabular}{@{}lrrrrrr@{}}
\toprule
                     & P-CQ  & R-CQ  & F1-CQ & P-REF & R-REF & F1-REF \\ \midrule
\multicolumn{7}{c}{\textit{ECD}}                                      \\ \midrule
Umberto-E3C-Pipeline & \textbf{86.81} & \textbf{77.32} & \textbf{79.48} & \textbf{92.43} & \textbf{83.28} & \textbf{86.49}  \\
Umberto-Base         & 82.11 & 73.62 & 77.21 & 87.45 & 77.81 & 81.89  \\
Word2Vec             & 76.21 & 67.55 & 71.24 & 81.67 & 72.35 & 76.73  \\
TF-IDF               & 50.11 & 41.20 & 42.43 & 52.54 & 44.81 & 45.45  \\
LDA                  & 27.74 & 34.04 & 28.48 & 19.19 & 27.57 & 29.90  \\ 
Embedding Match      & 68.29 & 58.44 & 59.21 & 70.59 & 61.52 & 62.09 \\ 
\midrule
\multicolumn{7}{c}{\textit{FEC}}                                      \\ \midrule
Umberto-E3C-Pipeline & \textbf{90.42} & \textbf{88.78} & \textbf{88.60} & \textbf{93.59} & \textbf{92.70} & \textbf{92.67}  \\
Umberto-Base         & 84.56 & 75.24 & 79.63 & 88.58 & 79.66 & 83.88  \\
Word2Vec             & 81.21 & 70.55 & 75.51 & 82.45 & 73.86 & 77.92  \\
TF-IDF               & 58.42 & 51.23 & 54.59 & 60.41 & 53.49 & 56.74  \\
LDA                  & 35.26 & 41.25 & 38.02 & 31.78 & 38.64 & 34.88 \\ 
Embedding Match      & 75.69 & 66.62 & 70.54 & 77.11 & 69.85 & 73.03 \\
\bottomrule
\end{tabular}
\footnotetext{Summary results of mapping to the disease labels defined by manual annotations on the test sets, computed for clinical questions (CQ) and referrals (REF). P=Precision- R=Recall- F1=F1-Score. The values of all metrics are expressed as percentages.}
\end{table}

\begin{table}[]
\centering
\caption{Summary evaluation on the appropriateness labels}
\label{tab:clusters-res-binary}
\begin{tabular}{@{}lrrrrrr@{}}
\toprule
                     & P-CQ  & R-CQ  & F1-CQ & P-REF & R-REF & F1-REF \\ \midrule
\multicolumn{7}{c}{\textit{ECD}}                                      \\ \midrule
Umberto-E3C-Pipeline & \textbf{89.24} & \textbf{88.27} & \textbf{88.64} & \textbf{93.58} & \textbf{91.52} & \textbf{92.51} \\
Umberto-Base         & 86.02 & 80.78 & 83.13 & 89.15 & 84.97 & 86.55  \\
Word2Vec             & 80.66 & 74.29 & 77.01 & 85.41 & 77.54 & 80.88  \\
TF-IDF               & 73.51 & 68.45 & 67.92 & 79.44 & 75.70 & 74.22  \\
LDA                  & 47.61 & 48.43 & 47.83 & 42.64 & 46.31 & 41.31  \\
Embedding Match      & 74.82 & 61.90 & 62.83 & 77.52 & 62.29 & 64.46 \\
\midrule
\multicolumn{7}{c}{\textit{FEC}}                                      \\ \midrule
Umberto-E3C-Pipeline & \textbf{92.05} & \textbf{90.96} & \textbf{90.88} & \textbf{94.66} & \textbf{93.96} & \textbf{94.05}  \\
Umberto-Base         & 87.41 & 80.55 & 83.84 & 89.62 & 82.16 & 85.73  \\
Word2Vec             & 83.24 & 74.15 & 78.43 & 85.77 & 75.63 & 80.38  \\
TF-IDF               & 67.11 & 57.96 & 62.20 & 68.14 & 60.85 & 64.29  \\
LDA                  & 38.96 & 45.62 & 42.03 & 35.41 & 42.93 & 38.81  \\ 
Embedding Match      & 80.07 & 69.92 & 74.20 & 81.15 & 73.92 & 77.15 \\

\bottomrule
\end{tabular}
\footnotetext{Summary results of mapping to the appropriate/inappropriate/likely (in)appropriate/undetermined categories on the manually labelled test sets, computed for clinical questions (CQ) and referrals (REF). P=Precision- R=Recall- F1=F1-Score. The values of all metrics are expressed as percentages.}
\end{table}

\subsection{Pipeline evaluation}

Table \ref{tab:clusters-res} presents the summary results of the evaluation performed on the manually annotated subsets, at the disease level, comparing different approaches. 
The pipeline with our fine-tuned Umberto embeddings demonstrates superior performance compared to alternative text representations and clustering algorithms, confirming the importance of contextual representations. The approach with direct matching between embeddings of clinical questions and embeddings of diseases from guidelines exhibits discrete results, but lower than our pipeline, confirming that the clustering procedure act as a useful smoothing step. The metrics are computed for both clinical questions (CQ) and referrals (REF). The referral-level scores are higher for all methods, except LDA, due to the influence of clinical questions that occur more frequently, which seem to be easier to cluster for them, being less detailed.

\begin{table}[]
\centering
\caption{Detailed results of the cluster mapping to the appropriateness labels for our pipeline}
\label{tab:detail_appr}
\begin{tabular}{@{}lrrrrrrrr@{}}
\toprule
               & P-CQ  & R-CQ   & F1-CQ & \multicolumn{1}{l}{\#CQ} & P-REF & R-REF  & F1-REF & \multicolumn{1}{l}{\# REF} \\ \midrule
\multicolumn{9}{c}{\textit{ECD}}                                                                                          \\ \midrule
Appropriate    & 98.55 & 93.15  & 95.77 & 365                      & 98.97 & 93.95  & 96.39  & 512                        \\
Inappropriate  & 91.29 & 81.88  & 86.33 & 320                      & 96.49 & 89.58  & 92.91  & 921                        \\
Likely inappr. & 43.28 & 89.23  & 58.29 & 65                       & 84.03 & 89.89  & 90.86  & 628                        \\
Undetermined   & 84.62 & 79.20  & 81.82 & 250                      & 87.22 & 83.33  & 85.23  & 426                        \\ \midrule
\multicolumn{9}{c}{\textit{FEC}}                                                                                          \\ \midrule
Appropriate    & 99.36 & 89.44  & 94.14 & 521                      & 99.88 & 96.96  & 98.40  & 2501                       \\
Inappropriate  & 95.35 & 87.23  & 91.11 & 47                       & 95.71 & 88.16  & 91.78  & 76                         \\
Likely appr.   & 98.27 & 87.18  & 92.39 & 195                      & 98.88 & 90.72  & 94.62  & 388                        \\
Undetermined   & 75.24 & 100.00 & 85.87 & 237                      & 84.19 & 100.00 & 91.42  & 591 \\                      
\bottomrule
\end{tabular}
\footnotetext{Detailed results of the cluster mapping to the appropriateness labels for our pipeline, on the manually labelled test sets, computed for clinical questions (CQ) and referrals (REF). P=Precision, R=Recall, F1=F1-Score. The values of all metrics are expressed as percentages.}
\end{table}

Table \ref{tab:clusters-res-binary} shows the same results achieved by grouping labels into the \textit{appropriate}, \textit{inappropriate}, \textit{likely (in)appropriate} and \textit{undetermined} categories.  These results show higher aggregate performance scores for ECD, compared to Table \ref{tab:clusters-res}, indicating that most errors occurred within the same appropriateness category rather than across categories. The results for FEC are also superior, but the improvement is limited. This suggests that distinguishing between appropriate and inappropriate referrals is simpler, while finer distinctions at the label level are more challenging. Table~\ref{tab:detail_appr} details the metric by appropriateness label, for our pipeline. At the clinical question level, it highlights issues with the likely inappropriate class for ECD, which corresponds directly to the \textit{varices} class at disease level, and it shows slightly lower results for the \textit{undetermined} class, but at the referrals level the only relevant difference is with the \textit{undetermined} class for ECD.
\begin{table}[tbh]
\centering
\caption{Detailed results of the cluster mapping for each disease label for our pipeline on ECD dataset}
\label{tab:cluster-det}
\begin{tabular}{p{2cm}rrrrrrrr}
\toprule
Category                     & P-CQ  & R-CQ  & F1-CQ   & \# CQ & P-REF & R-REF & F1-REF & \# REF \\
\midrule
Claudicatio          & 83.33  & 45.45 & 58.82 & 11          & 90.00        & 56.25       & 69.23      & 16           \\
Pain                 & 85.00  & 91.07 & 87.93 & 56          & 88.61        & 93.33       & 90.91      & 75           \\
Edema                & 85.47  & 96.15 & 90.50 & 104         & 89.51        & 97.32       & 93.25      & 149          \\
Venous insuff. & 94.90  & 74.50 & 83.47 & 200         & 98.06        & 88.09       & 92.81      & 747          \\
Paresthesis          & 100.00 & 81.25 & 89.66 & 16          & 100.00       & 88.00       & 93.62      & 25           \\
Saphenectomy         & 100.00 & 92.59 & 96.15 & 27          & 100.00       & 93.94       & 96.88      & 33           \\
(Trombo) Phlebitis    & 98.70  & 93.83 & 96.20 & 81          & 99.09        & 93.16       & 96.04      & 117          \\
Thrombosis           & 96.24  & 94.04 & 95.13 & 218         & 96.99        & 94.77       & 95.87      & 306          \\
Ulcer                & 95.83  & 92.00 & 93.88 & 25          & 97.44        & 92.68       & 95.00      & 41           \\
Varices              & 43.28  & 89.23 & 58.29 & 65          & 84.03        & 98.89       & 90.86      & 628          \\
Varices surgery      & 100.00 & 42.86 & 60.00 & 14          & 100.00       & 46.67       & 63.64      & 15           \\
Vasculopathy         & 82.14  & 44.23 & 57.50 & 52          & 95.33        & 68.92       & 80.00      & 148 \\
Other                & 63.57  & 67.94 & 65.68 & 131         & 62.56        & 70.59       & 66.33      & 187          \\
\bottomrule
\end{tabular}
\footnotetext{Detailed results of the cluster mapping for each disease for our pipeline, on the manually labelled test set for ECD, computed for clinical questions (CQ) and referrals (REF). Labels with support $< 10$ are grouped with \textit{Other}. P=Precision, R=Recall, F1=F1-Score. The values of all metrics are expressed as percentages.}
\end{table}

\begin{table}[tbh]
\centering
\caption{Detailed results of the cluster mapping for each disease label for our pipeline on FEC dataset}
\label{tab:colon-classes-res}
\begin{tabular}{p{2cm}rrrrrrrr}
\toprule
Category & P-CQ & R-CQ & F1-CQ & \# CQ & P-REF & R-RE & F1-REF & \# REF \\ \midrule
Bleeding                           & 100.00 & 77.91 & 87.59 & 163 & 100.00 & 97.31 & 98.64 & 1748 \\
Polyps                             & 99.35  & 97.47 & 98.40 & 158 & 99.71  & 98.55 & 99.12 & 344  \\
Family history / prevention        & 97.56  & 86.33 & 91.60 & 139 & 98.57  & 91.39 & 94.85 & 302  \\
Colon cancer                       & 96.84  & 85.98 & 91.09 & 107 & 98.24  & 89.30 & 93.56 & 187  \\
Abdominal pain                     & 88.57  & 89.42 & 89.00 & 104 & 93.66  & 89.72 & 91.65 & 214  \\
Anaemia                            & 92.75  & 98.46 & 95.52 & 65  & 95.73  & 97.39 & 96.55 & 115  \\
Chronic inflammatory bowel disease & 100.00 & 89.29 & 94.34 & 56  & 100.00 & 88.37 & 93.83 & 86   \\
Constipation                       & 78.38  & 80.56 & 79.45 & 36  & 91.61  & 94.24 & 92.91 & 139  \\
Diverticulosis                     & 90.00  & 81.82 & 85.71 & 33  & 86.67  & 81.25 & 83.87 & 48   \\
Altered bowel                      & 84.85  & 93.33 & 88.89 & 30  & 92.54  & 95.38 & 93.94 & 65   \\
Chronic diarrhoea                  & 100.00 & 82.14 & 90.20 & 28  & 100.00 & 93.46 & 96.62 & 107  \\
Diverticulitis                     & 100.00 & 92.86 & 96.30 & 14  & 100.00 & 89.29 & 94.34 & 28   \\
Other                              & 47.14  & 98.51 & 63.77 & 67  & 59.93  & 99.42 & 74.78 & 173  \\ \bottomrule
\end{tabular}
\footnotetext{
Detailed results of the cluster mapping for each disease for our pipeline, on the manually labelled test set for FEC, computed for clinical questions (CQ) and referrals (REF). Labels with support $< 10$ are grouped with \textit{Other}. P=Precision, R=Recall, F1=F1-Score. The values of all metrics are expressed as percentages.
}
\end{table}

Tables \ref{tab:cluster-det} and \ref{tab:colon-classes-res} detail metrics by disease for our pipeline. Most diseases achieve \textit{F1-CQ} $> 80\%$, with many exceeding $> 90\%$. Lower scores are observed for some ECD categories such as \textit{varices}, \textit{varices surgery}, \textit{vasculopathy}, \textit{claudicatio} and \textit{other}, while for FEC only \textit{other} exhibits poor precision.
At the referral level (\textit{F1-REF}), only \textit{claudicatio}, \textit{varices surgery} and \textit{other} achieve scores below $80 \%$ for ECD, while for FEC only \textit{other} is below $80 \%$. 

\begin{figure}
    \centering
    \includegraphics[width=\textwidth]{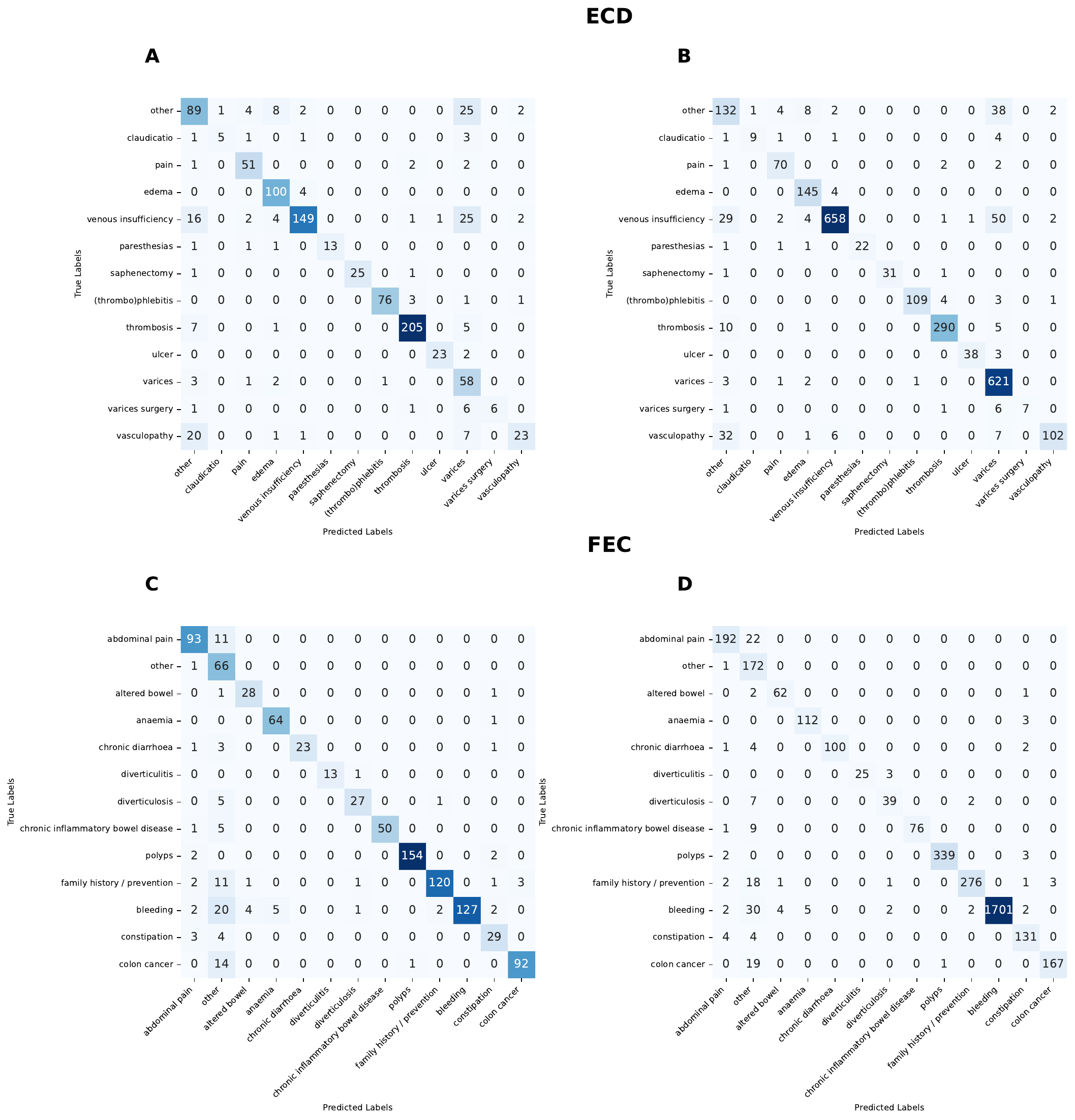}
    \caption{Confusion matrix of the clustering assignment to the labels on the manually annotated test sets, for ECD (A, B) and FEC (C,D) computed considering the number of clinical questions (A, C) and the number of referrals (B, D) }
    \label{fig:conf-matrix}
\end{figure}

The confusion matrices shown in Figure \ref{fig:conf-matrix} allow for a more detailed analysis of misclassification errors.
For ECD, errors in \textit{varices surgery} are often due to misclassifications as \textit{varices}, a strictly related cluster. Regarding \textit{varices}, its precision is mainly affected by misclassifications from \textit{venous insufficiency} and \textit{other}. Looking at the specific clinical questions manually labelled as \textit{venous insufficiency} and classified as \textit{varices}, we notice that some present both diagnoses, making this misclassification reasonable. For the remaining ECD classes, misclassified samples fall mainly into the \textit{other} group, affecting its precision. 
For FEC, the majority of misclassification errors are wrong assignments to the \textit{other} class. A few are wrong assignments to \textit{abdominal pain}, which are in multiple cases due to the mention of this symptom along with others that were more relevant.
A detailed list of all identified clusters, together with their mapping, is reported in Additional File 1, Section C.

\begin{figure}
    \centering
    \includegraphics[width=.65\linewidth]{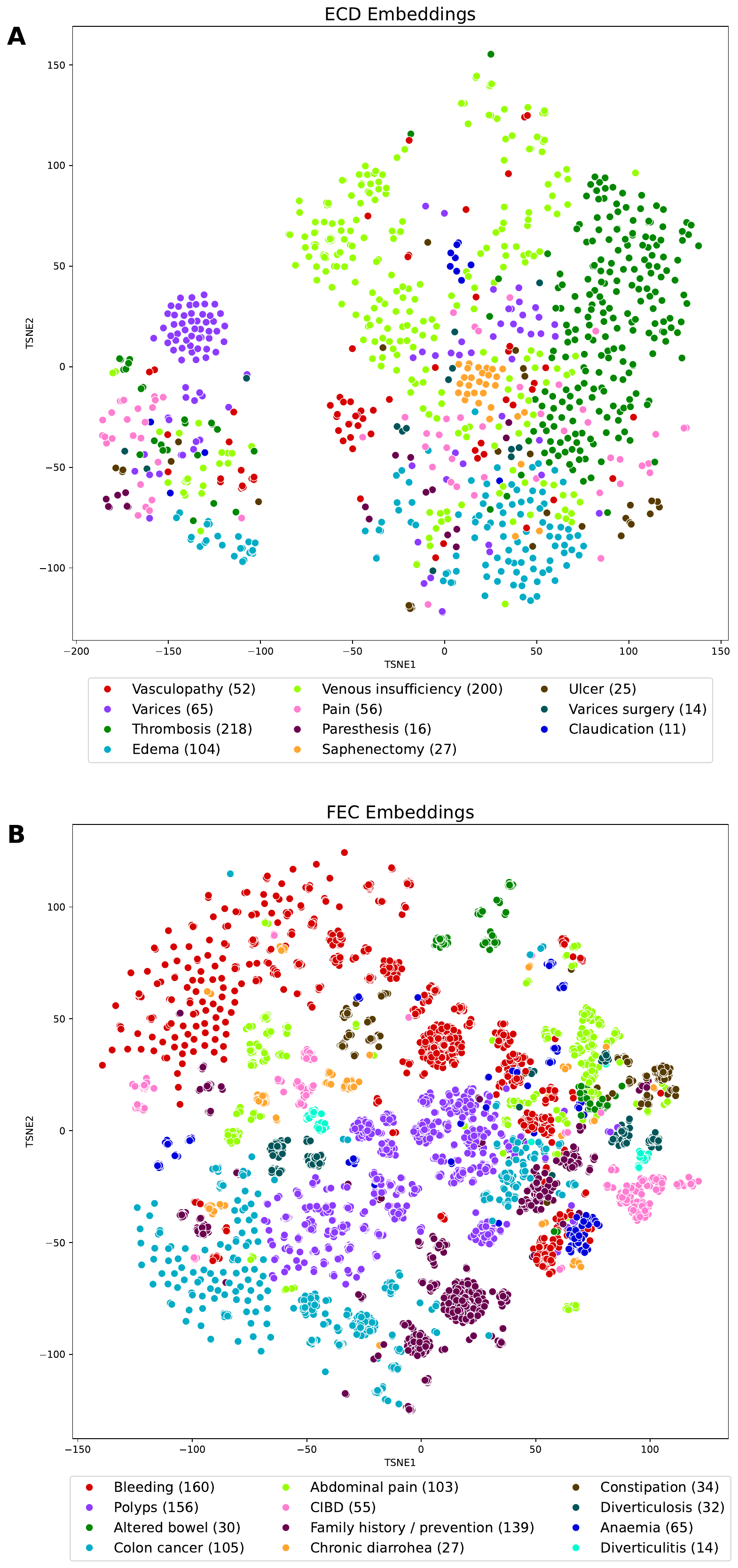}
    \caption{2D t-SNE projections of embeddings of clinical questions from the validation sets of ECD (A) and FEC (B) with their labels, excluding the \textit{Other} class}
    \label{fig:embeddings}
\end{figure}

Figure~\ref{fig:embeddings} displays t-SNE 2D projections of the embedding vectors of the clinical questions in the two labelled subsets, showing that for most classes there are one or more distinct regions of the 2D space that tend to concentrate the corresponding clinical questions.

\subsection{Analysis of the entire dataset}
\begin{table}[hbt!]
\centering
\caption{Summary of referrals' clusters for ECD with appropriateness labels}
\label{tab:clusters-appr}
\begin{tabular}{lrr}
\toprule
Cluster Label & \multicolumn{1}{c}{\# REF} & \multicolumn{1}{c}{\%} \\
\rowcolor[HTML]{EA9999}
\midrule 
venous insufficiency      & 102,339 & 20.63 \\
\rowcolor[HTML]{B7E1CD} 
thrombosis                & 98,821  & 19.92 \\
\rowcolor[HTML]{FFE599} 
varices                   & 71,269  & 14.37 \\
\rowcolor[HTML]{EA9999} 
edema                     & 61,822  & 12.46 \\
\rowcolor[HTML]{B7E1CD} 
(thrombo)phlebitis        & 49,537  & 9.99  \\
\rowcolor[HTML]{EFEFEF} 
pain                      & 21,131  & 4.26  \\
\rowcolor[HTML]{EFEFEF} 
vasculopathy               & 11,244  & 2.27  \\
\rowcolor[HTML]{B7E1CD} 
saphenectomy              & 8,290   & 1.67  \\
\rowcolor[HTML]{B7E1CD} 
varices surgery           & 7,126   & 1.44  \\
\rowcolor[HTML]{EFEFEF} 
diabetes                  & 6,583   & 1.33  \\
\rowcolor[HTML]{B7E1CD} 
ulcer                     & 6,272   & 1.26  \\
\rowcolor[HTML]{EA9999} 
paresthesias              & 4,856   & 0.98  \\
\rowcolor[HTML]{EFEFEF} 
claudicatio               & 3,570   & 0.72  \\
\rowcolor[HTML]{EFEFEF} 
dimer                     & 1,601   & 0.32  \\
\rowcolor[HTML]{EFEFEF} 
swelling                  & 1,238   & 0.25  \\
\rowcolor[HTML]{EFEFEF} 
respiratory insufficiency & 1,060   & 0.21  \\
\rowcolor[HTML]{EFEFEF} 
heavy legs                & 705    & 0.14  \\
\rowcolor[HTML]{EFEFEF} 
ischaemia                 & 617    & 0.12  \\
\rowcolor[HTML]{EFEFEF} 
stent                     & 612    & 0.12  \\
\rowcolor[HTML]{EFEFEF} 
dermatitis                & 474    & 0.10  \\
\rowcolor[HTML]{EFEFEF} 
hypercolesterolaemia      & 416    & 0.08  \\
\rowcolor[HTML]{EFEFEF} 
pregnancy                 & 225    & 0.05  \\
\rowcolor[HTML]{EFEFEF} 
dialisis                  & 222    & 0.04  \\
\rowcolor[HTML]{B7E1CD} 
surgery                   & 198    & 0.04  \\
\rowcolor[HTML]{EFEFEF} 
smoker                    & 193    & 0.04  \\
\rowcolor[HTML]{EFEFEF} 
dyspnoea                  & 171    & 0.03  \\
\rowcolor[HTML]{EFEFEF} 
tingling                  & 170    & 0.03  \\
\rowcolor[HTML]{EFEFEF} 
cold legs                 & 147    & 0.03  \\
\rowcolor[HTML]{EFEFEF} 
hypoesthesia              & 80     & 0.02  \\
\rowcolor[HTML]{EFEFEF} 
metabolic syndrome        & 48     & 0.01  \\
\rowcolor[HTML]{EFEFEF} 
bleeding                  & 29     & 0.01  \\
\rowcolor[HTML]{EFEFEF} 
other                     & 34,917  & 7.04 \\ 
\bottomrule
\end{tabular}
\footnotetext{
Summary of referrals' clusters for ECD, mapped to all labels defined in manual annotations. Green indicates appropriate clusters, red inappropriate clusters, yellow clusters for which the information might not be sufficient to define the appropriateness but is more likely inappropriate, and light grey those not mentioned in the guidelines.
}
\end{table}

\begin{table}[tbh]
\centering
\caption{Summary of referrals' clusters for FEC with appropriateness labels}
\label{tab:gastro-appr-res}
\begin{tabular}{lrr}
\toprule
Cluster Label &                           \# REF & \multicolumn{1}{c}{\%} \\
\rowcolor[HTML]{B7E1CD} 
\midrule
polyps                                & 81,111 & 19.88 \\
\rowcolor[HTML]{B7E1CD} 
occult blood                          & 76,487 & 18.75 \\
\rowcolor[HTML]{FFE599} 
colon cancer familiarity / prevention & 43,615 & 10.69 \\
\rowcolor[HTML]{EFEFEF} 
abdominal pain                          & 31,316 & 7.68  \\
\rowcolor[HTML]{B7E1CD} 
colon cancer                          & 30,434 & 7.46  \\
\rowcolor[HTML]{B7E1CD} 
anaemia                               & 23,133 & 5.67  \\
\rowcolor[HTML]{FFE599} 
chronic inflammatory bowel disease    & 21,470 & 5.26  \\
\rowcolor[HTML]{EFEFEF} 
constipation                          & 15,109 & 3.70  \\
\rowcolor[HTML]{EA9999} 
diverticulosis                        & 15,058 & 3.69 \\
\rowcolor[HTML]{B7E1CD} 
chronic diarrhoea                     & 13,533 & 3.32  \\
\rowcolor[HTML]{EFEFEF} 
altered bowel                          & 10,933 & 2.68  \\
\rowcolor[HTML]{EFEFEF} 
haemorrhoidal disease                 & 5,379  & 1.32  \\
\rowcolor[HTML]{EA9999} 
diverticulitis                        & 4,669  & 1.14  \\
\rowcolor[HTML]{B7E1CD} 
suspected colic lesion detected by other investigation technique & 3671                                & 0.90                                \\
\rowcolor[HTML]{EA9999} 
irritable bowel syndrome                       & 2,023  & 0.50  \\
\rowcolor[HTML]{EFEFEF} 
surgery                       & 797   & 0.20  \\
\rowcolor[HTML]{EFEFEF} 
lynch syndrome                        & 787   & 0.19  \\
\rowcolor[HTML]{EFEFEF} 
other                                 & 28,424 & 6.98 \\
\bottomrule
\end{tabular}
\footnotetext{
Summary of referrals' clusters for FEC, mapped to all labels defined in manual annotations. Green indicates appropriate clusters, red inappropriate clusters, yellow clusters for which the information might not be sufficient to define the appropriateness but is more likely appropriate, and light grey those not mentioned in the guidelines.
}
\end{table}

Tables \ref{tab:clusters-appr} and \ref{tab:gastro-appr-res} extend the analysis to the entire ECD and FEC datasets, highlighting the appropriateness labels for each cluster. We can observe that the distribution of the reasons for the referrals is unbalanced, with 75\% of referrals covered by the first five reasons in ECD and by the first seven reasons in FEC. 
In ECD, \textit{varices} cluster was assigned to \textit{likely inappropriate} since this diagnosis is inappropriate unless it is associated with surgery, an element that is often difficult to exclude, even for a human reader. 
In FEC, \textit{colon cancer familiarity / prevention} and \textit{chronic inflammatory bowel disease} were instead marked as \textit{likely appropriate} since guidelines state that physicians should carefully consider the patient history and condition to establish if and when they should be performed. Specific conditions mentioned in the guidelines — such as requiring ten years from diagnosis or significant colon involvement — were largely absent from referral texts.

Summarizing Table \ref{tab:clusters-appr}, for ECD, 34.32\% of referrals appear to be appropriate, while 34.07\% seems to be not appropriate. The remaining 31.61\% referrals can be further divided into 14.37\% more likely to be not appropriate, 10.20\% that are not mentioned on the guidelines, and 7.04\% that cannot be clustered or cannot be mapped to the labels defined by manual annotations. 
For FEC, 55.98\% of referrals appear to be appropriate, while only 5.33\% seems to be not appropriate. There are then 15.95\% more likely to be appropriate, 15.77\% that are not mentioned on the guidelines, and 6.98\% that cannot be clustered or cannot be mapped to the labels defined by manual annotations.
Notably, in both datasets the \textit{other} group is limited, while there are clusters for many symptoms and diseases not mentioned in the guidelines, especially for ECD. 
Although individually rare, these uncovered reasons accounted for a non-negligible proportion of overall referrals, suggesting the need for a broader guideline scope. Additionally, we observed ambiguity in referrals citing varices, where physicians often omitted whether surgical intervention was required. 
In contrast, FEC referrals showed higher levels of appropriateness, though some inappropriate clusters were identified. 

\begin{figure}
    \centering
    \includegraphics[height=0.9\textheight]{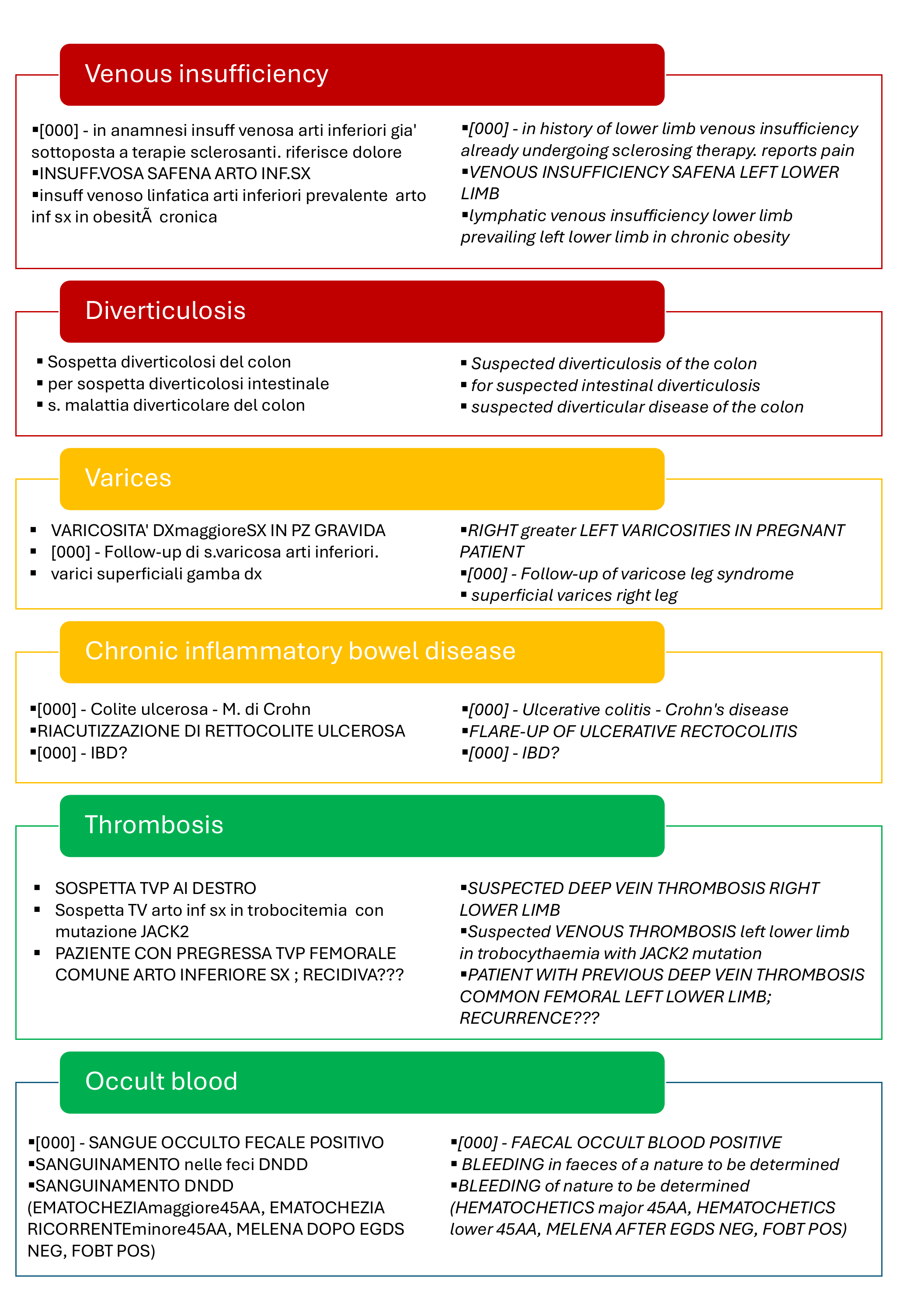}
    \caption{Examples of clustered clinical questions. English translations do not completely reflect the abbreviations and typos of the original Italian texts.}
    \label{fig:examples-clustered}
\end{figure}

Figure \ref{fig:examples-clustered} provides some examples, along with their English translation, of clustered clinical questions.

\begin{figure}
    \centering
    \includegraphics[height=0.9\textheight]{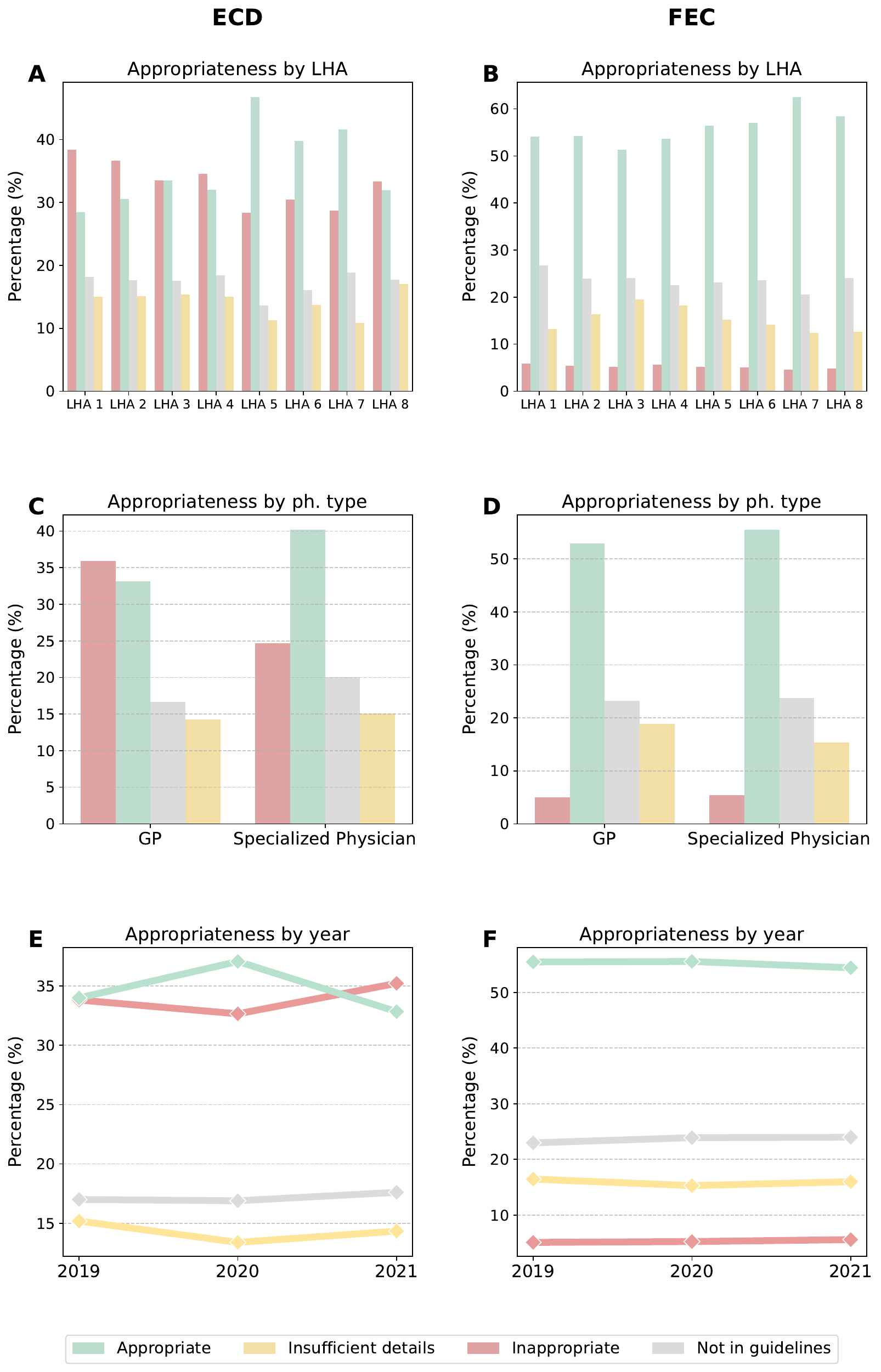}
    \caption{Prescription appropriateness stratified by  LHA (A, B), type of prescribing physicians (C, D) and year (E, F) for ECD (A,C,E) and FEC (B,D,F).}
    \label{fig:summary-strata}
\end{figure}

Figure \ref{fig:summary-strata} reports the level of appropriateness stratified by LHA, type of prescribing physicians and year, providing additional insights into the reasons for inappropriate prescriptions. 
This stratified analysis reveals significant geographical variability for both examinations and especially for ECD, suggesting that targeted interventions are needed in specific Local Health Authorities. General practitioners exhibited higher rates of inappropriate ECD referrals compared to specialist physicians. This might have multiple reasons, including a larger patient base, more frequent encounters with less severe cases, and a lower knowledge and/or agreement with guidelines~\cite{piccoliori2024role, lane2013management}. For FEC, the rate of inappropriateness is similar but specialized physicians have more appropriate referrals.
An important trend emerges when analyzing the impact of COVID-19 on referral practices. During the early pandemic phase in 2020, inappropriate ECD referrals dropped significantly, likely due to resource constraints prompting physicians to prioritize urgent cases. This pattern is consistent with existing literature, which reports a general reduction in non-urgent medical procedures and a shift toward more conservative referral behaviour during the pandemic~\cite{maida2020impact, di2020impact, moynihan2020covid, coccia2021impact}. However, this decline was temporary, with inappropriate referrals rebounding in 2021 as healthcare services normalized. A similar, though less pronounced, trend was observed for FEC, possibly due to the already limited number of inappropriate referrals for this examination.
\section{Discussion}
\label{sec:discussion}
In this study, we proposed a methodological pipeline leveraging NLP techniques to assess the appropriateness of diagnostic referrals in the Italian NHS. We illstrated its use through two case studies on \textit{venous echocolordoppler of the lower limbs} in angiology and \textit{flexible endoscope colonoscopy} in gastroenterology. 
By applying an unsupervised approach, we uncovered key trends in referral motivations and guideline adherence, offering the first large-scale assessment of this kind in Italy.

Our work presents several strengths. The pipeline operates in an unsupervised setting, without the need for labelled data, which are costly and time-consuming to produce. This makes it applicable to any examination with established guidelines defined in terms of appropriate and inappropriate reasons for referrals. It is also designed to run on large datasets without requiring high computational resources, lowering the barrier for adoption. In addition to quantifying appropriateness, the pipeline offers a detailed picture of referral motivations, thereby revealing reasons not explicitly covered by existing guidelines.

We demonstrated these capabilities on two population-scale datasets, covering an entire Italian region with 10 million inhabitants for three years. Validation on manually annotated subsets confirmed the robustness of our pipeline and demonstrated that contextual embedding models significantly outperform simpler clustering techniques in capturing the semantic nuances of referral texts. We observed consistently high performances at the appropriate/inappropriate level and for most diseases/symptoms, with only two clusters in ECD with lower recall and one cluster in FEC with lower precision. The excellent results on two different examinations, one of which was not used during pipeline development, confirmed the generalizability of our methodology.

The large-scale analysis enabled, for the first time, a comprehensive view of referral motivations for these examinations. This is a relevant result for public health authorities, given the absence of structured recording of this information in the Italian NHS. The identification of undocumented reasons for referrals highlighted potential gaps in the existing guidelines and the stratified analyses further revealed geographical and temporal patterns and variations across physician types, consistent with findings in the literature and suggestive of targeted interventions to reduce inappropriate referrals. Importantly, the findings directly supported the Lombardy Region in issuing a new resolution extending and reinforcing referral guidelines. This represents one of the first documented cases in Italy where an AI-based analysis of routine healthcare data directly shaped health policy.

This study is a relevant example of a real-world, large-scale application of artificial intelligence in healthcare. Despite a growing body of research, the adoption of AI-based solutions in clinical practice faces significant challenges~\cite{carini2024tribulations} and remains largely dominated by diagnostic tools in medical imaging~\cite{rajpurkar2022ai, bizzo2023addressing}, particularly in Italy~\cite{ardito2025adoption}. Our work points to a complementary direction: leveraging large-scale textual data to monitor referral practices. In this way, it illustrates how AI can extend its role beyond diagnosis to support healthcare governance and inform evidence-based policy.

Despite its strengths, this study has some limitations. Manual annotation was limited to a subset of the data due to time constraints, and performed by a single annotator, limiting the possibility of inter-annotator agreement analysis. Our approach also assigns a single primary reason per referral, potentially overlooking multi-reason referrals, which could be better addressed through fuzzy clustering techniques. Moreover, the lack of access to patient histories limited our ability to fully assess clinical appropriateness in certain cases, such as referrals for \textit{varices}, where prior surgical indication is relevant. 
It is important to clarify that our pipeline does not aim to assess the appropriateness of individual referrals. Instead, our approach provides a systematic, automated, and large-scale evaluation of prescription practices at the regional level, offering valuable insights for policymakers and healthcare administrators. 

In future work, we aim to extend the validation of the pipeline to other examinations and broader settings, including larger annotated datasets and other Italian regions, to further support the adoption of data-driven tools for evaluating referral practices and improving healthcare delivery.
\section{Conclusions}
\label{sec:conclusions}
This study proposed a methodological pipeline for assessing the level of appropriateness of Italian referrals.
We presented two case studies on \textit{venous echocolordoppler of the lower limbs} and \textit{flexible endoscope colonoscopy} in the Lombardy Region, demonstrating excellent performance on manually annotated test sets and deriving relevant insights for public health authorities. The outcome of this analysis allowed the Lombardy Region to reinforce existing guidelines for physicians, with the aim of reducing the number of inappropriate referrals. 
This work is an example of how artificial intelligence can be deployed on large, real-world administrative datasets, highlighting the potential of AI to support healthcare systems in monitoring practices, informing policy, and ultimately alleviating significant burdens on the Italian NHS by lowering waiting times and enhancing overall effectiveness of care.

\backmatter

\bmhead{Supplementary information}
\begin{itemize}
    \item Name: Additional file 1  \\
Format: pdf \\
Title: Supplementary Material \\
Description: contains more details about the manual annotation process, hyperparameters selection and additional results
\end{itemize}

\bmhead{List of abbreviations}
\begin{enumerate}
    \item AI: Artificial Intelligence
    \item BERT: Bidirectional Encoder Representation for Transformers
    \item CIBD: Chronic Inflammatory Bowel Disease
    \item CQ: Clinical Question   
    \item DGR: Decreto della Giunta Regionale (Regional Law)
    \item ECD: venous EcoColorDoppler of the lower limbs
    \item EHR: Electronic Health Record
    \item FEC: Flexible Endoscope Colonscopy with possible biopsies
    \item GP: General Practitioner
    \item ICD: International Classification of Diseases
    \item IQR: InterQuartile Range
    \item LDA: Latent Dirichlet Allocation
    \item ATS/LHA: Agenzia di Tutela della Salute / Local Health Authority
    \item NHS: National Health Service
    \item NLP: Natural Language Processing
    \item REF: Referral    
    \item TF-IDF: Term Frequency - Inverse Document Frequency
\end{enumerate}

\section*{Declarations}

\subsection*{Ethics approval and Consent to participate}
The need for ethics approval is deemed unnecessary according to national regulations. In particular, according to the rules from the Italian Medicines Agency (available at: \url{https://www.gazzettaufficiale.it/eli/id/2024/08/20/24A04320/sg}), retrospective studies using administrative databases do not require Ethics Committee protocol approval. 

\noindent The need for consent to participate is deemed unnecessary according to national regulations. In particular, see the rules of the Italian Privacy Authority available at \url{https://www.garanteprivacy.it/web/guest/home/docweb/-/docweb-display/docweb/9124510} and the European Union regulation n. 2016/679 (GDPR) Art. 1 c. 1 (\url{https://gdpr.eu/article-1-subject-matter-and-objectives-overview/}), Art. 4 c.1 (\url{https://gdpr.eu/article-4-definitions/}) and Recital 26 (\url{https://gdpr-info.eu/recitals/no-26/}).

\subsection*{Consent for publication}
Not applicable.

\subsection*{Availability of data and materials}
The data that support the findings of this study are available from Regione Lombardia but restrictions apply to the availability of these data, which were used under license for the current study, and so are not publicly available. Data are however available from the authors upon reasonable request and with permission of Regione Lombardia.

\subsection*{Code availability}
The code is available at \url{https://github.com/vittot/ReferralsAppropriateness}

\subsection*{Competing interests}
The authors declare that they have no competing interests.

\subsection*{Funding}
No fundings supported this research.

\subsection*{Authors' contributions}
VT: analysis and interpretation of data, development of the methodology, drafting of the manuscript
AB: concept and design of the study, acquisition, analysis and interpretation of data
ME: concept and design of the study, acquisition, analysis and interpretation of data
OL: concept and design of the study, acquisition, analysis and interpretation of data
FI: concept and design of the study, analysis and interpretation of data, critical revision of the manuscript, supervision.
All authors read and approved the final manuscript.

\subsection*{Acknowledgements}
The present research is part of the activities of “Dipartimento di Eccellenza 2023-2027”.
We thank the Welfare Directorate of Regione Lombardia (Agreement between Politecnico di Milano and Regione Lombardia DGR n. XI/5301-27/09/2021) and, in particular, the working group on waiting list monitoring. 
We thank Dr. Paola Colombo for the consultation regarding the reasons for angiology referrals and Dr. Marco Soncini for gastroenterology referrals.



\bibliography{sn-bibliography}

\clearpage

\renewcommand\thefigure{\thesection.\arabic{figure}}    
\renewcommand{\thetable}{\thesection.\arabic{table}}

\appendix

\pagenumbering{Roman}

\section{Manual annotations}
\label{app:annotations}
\setcounter{figure}{0}  
\setcounter{table}{0}
A validation set of 1,000 clinical questions was manually annotated for the main disease/symptom determining the referrals, for both ECD and FEC datasets. 
The existing guidelines from the Lombardy Region~\cite{dgrAngio, dgrGastro} were used as a reference, but labels were not forced to be one of the categories mentioned in the guidelines as appropriate or inappropriate. If none of the categories mentioned in the guidelines fit the reason for the referral, a new category representing the main disease or symptom mentioned in the referral was created. The \textit{other} category was used only when it was impossible to identify a symptom or a disease determining the referral (e.g., ``\textit{in-depth examination to actually establish whether there is an ongoing pathology given the symptoms present}" or ``\textit{gfwhdfg}").

In case of multiple diseases or symptoms, the following priority order was defined:
\begin{enumerate}
    \item Reasons mentioned as appropriate in the guidelines
    \item Reasons mentioned as inappropriate among the guidelines
    \item Reasons not mentioned in the guidelines
\end{enumerate}
If considering these priority levels, there were still multiple diseases/symptoms, the following priority order was defined:
\begin{enumerate}
    \item If there is one symptom/disease which looks like the leading cause, prefer it
    \item If there are multiple symptoms/diseases mentioned in the guidelines, choose the first one to be mentioned in the text
    \item If there are multiple symptoms/diseases not mentioned in the guidelines, prefer those for which a category was already created, otherwise the first one to be mentioned in the text
\end{enumerate}
This procedure, after four iterations, including consultation with specialist physicians, defined the labels reported in Tables \ref{tab:manual-ann-ecd-details} and \ref{tab:manual-ann-gastro-details}. After review of these labels, some of them were merged:
\begin{itemize}
    \item \textit{bleeding} and \textit{occult blood} since bleeding in this context is always an occult bleeding
    \item \textit{colitis}, \textit{chron's disease}, \textit{ulcerative rectocolitis},  \textit{chronic inflammatory bowel disease}, under the latter which is a category comprehensive of all these diseases
\end{itemize}

\begin{table}
\caption{Labels derived by manual annotations on the validation set of 1,000 clinical questions for ECD.}
\label{tab:manual-ann-ecd-details}
\scriptsize
\begin{tabular}{@{}lrr@{}}
\toprule
Label                   & \# CQ & \# REF \\ 
\midrule
Thrombosis              & 218     & 306                        \\
Venous insufficiency    & 200     & 747                        \\
Edema                   & 104     & 149                        \\
(Thrombo)phlebitis      & 81      & 117                        \\
Varices                 & 65      & 628                        \\
Pain                    & 56      & 75                         \\
Vasculopaty             & 52      & 148                        \\
Safenectomy             & 27      & 33                         \\
Surgery                 & 26      & 26                         \\
Ulcer                   & 25      & 41                         \\
Paresthesis             & 16      & 25                         \\
Varices surgery         & 14      & 15                         \\
Other                   & 13      & 16                         \\
Claudicatio             & 11      & 16                         \\
Swelling                & 9       & 10                         \\
Transplanctation        & 8       & 8                          \\
Teleangectasias         & 7       & 7                          \\
Cardiopathy             & 6       & 15                         \\
Dermatitis              & 6       & 12                         \\
Dimer                   & 6       & 6                          \\
Ischaemia               & 5       & 5                          \\
Diabetes                & 4       & 5                          \\
Prosthesis              & 4       & 4                          \\
Cancer                  & 3       & 5                          \\
Heavy Legs              & 3       & 4                          \\
Hypertension            & 3       & 6                          \\
Stent                   & 3       & 2                          \\
Arthritis               & 2       & 3                          \\
Calcification           & 2       & 2                          \\
Cold legs               & 2       & 2                          \\
Dyslipidaemia           & 2       & 2                          \\
Tingling                & 2       & 3                          \\
Dialysis                & 1       & 1                          \\
Dyspnoea                & 1       & 1                          \\
Embolism                & 1       & 2                          \\
Haemorrhage             & 1       & 1                          \\
Hypercholesterolaemia   & 1       & 3                          \\
Hypertrophy             & 1       & 1                          \\
Hyposthesia             & 1       & 1                          \\
Infection               & 1       & 1                          \\
Lymphangitis            & 1       & 1                          \\
Metabolic syndrome      & 1       & 1                          \\
Multiple sclerosis      & 1       & 1                          \\
Neuropathy              & 1       & 1                          \\
Pregnancy               & 1       & 1                          \\
Smoker                  & 1       & 1                          \\
Vascular encephalopathy & 1       & 1                          \\ 
\bottomrule
\end{tabular}
\end{table}
\begin{table}
\caption{Labels derived by manual annotations on the validation set of 1,000 clinical questions for FEC}
\label{tab:manual-ann-gastro-details}
\scriptsize
\begin{tabular}{lrr}
\toprule
\multicolumn{1}{l}{Label}          & \multicolumn{1}{l}{\# CQ} & \multicolumn{1}{l}{\# REF} \\ \midrule
Polyps                             & 158                       & 344                        \\
Bleeding                           & 127                       & 959                        \\
Colon cancer                       & 107                       & 187                        \\
Abdominalgia                       & 104                       & 214                        \\
Colon cancer familiarity           & 82                        & 141                        \\
Anaemia                            & 65                        & 115                        \\
Prevention                         & 57                        & 161                        \\
Occult blood                       & 36                        & 789                        \\
Constipation                       & 36                        & 139                        \\
Diverticulosis                     & 33                        & 48                         \\
Altered bowel                      & 30                        & 65                         \\
Colitis                            & 28                        & 40                         \\
Chronic diarrhoea                  & 28                        & 107                        \\
Diverticulitis                     & 14                        & 28                         \\
Crohn's disease                    & 12                        & 16                         \\
Haemorrhoidal disease              & 9                         & 30                         \\
Chronic inflammatory bowel disease & 8                         & 15                         \\
Ulcerative rectocolitis            & 8                         & 15                         \\
Surgery                            & 5                         & 6                          \\
Intestinal obstruction             & 5                         & 10                         \\
Dyspepsia                          & 4                         & 5                          \\
Weight loss                        & 3                         & 8                          \\
Irritable bowel syndrome           & 3                         & 6                          \\
Lymphoma                           & 3                         & 4                          \\
Transplant                         & 3                         & 15                         \\
Other                              & 2                         & 2                          \\
Colonpathy                         & 2                         & 7                          \\
Enteritis                          & 2                         & 3                          \\
Proctitis                          & 2                         & 4                          \\
Rectal prolapse                    & 2                         & 12                         \\
Lynch syndrome                     & 2                         & 2                          \\
Angiodysplasia                     & 1                         & 1                          \\
Appendicitis                       & 1                         & 1                          \\
Celiac disease                     & 1                         & 1                          \\
Intestinal loops distension        & 1                         & 1                          \\
Hepatopathy                        & 1                         & 1                          \\
Protruding formation               & 1                         & 1                          \\
Gastritis                          & 1                         & 2                          \\
Ileitis                            & 1                         & 1                          \\
Hypercholesterolaemia              & 1                         & 1                          \\
Splenic flexure thickening         & 1                         & 3                          \\
Rectal wall thickening             & 1                         & 2                          \\
Sigma wall thickening              & 1                         & 1                          \\
Behcet's disease                   & 1                         & 1                          \\
Liver disease                      & 1                         & 1                          \\
Pancreatitis                       & 1                         & 1                          \\
Proctalgia                         & 1                         & 3                          \\
Sigmoiditis                        & 1                         & 1                          \\
Suspected colic lesion                   & 1                         & 33                         \\
Splenic fissure stenosis           & 1                         & 1                          \\
Rectal tenesmus                    & 1                         & 2                          \\ \bottomrule
\end{tabular}
\end{table}

\section{Pre-processing results}
\label{app:preproc}
\setcounter{figure}{0}  
\setcounter{table}{0}
Table \ref{tab:abbr} reports the list of the 50 most frequent abbreviations that were expanded by our pre-processing, while Table \ref{tab:typos} contains an example list of 50 misspelt words that were detected and corrected during our pre-processing. Table \ref{tab:res-wo-preproc} reports the summary results on disease labels without applying pre-processing. 

\begin{table}[h!]
\centering
\caption{Most frequent abbreviations in clinical questions for echocolordoppler of the venous limbs, along with their expansions in Italian and English.}
\label{tab:abbr}
\scriptsize
\begin{tabular}{@{}lrll@{}}
\toprule
Abbr & Freq & Expansion                          & Expansion {[}ENG{]}                 \\ \midrule
dx                       & 41942                    & destra                       & right                         \\
sx                       & 35400                    & sinistra                     & left                          \\
tvp                      & 35392                    & trombosi venosa profonda     & deep vein thrombosis          \\
inf                      & 26644                    & inferiore                    & inferior                      \\
sin                      & 8490                     & sinistra                     & left                          \\
pz                       & 6100                     & paziente                     & patient                       \\
ndd                      & 4555                     & di natura da determinare     & of a nature to be determined  \\
aa                       & 4079                     & arti                         & limbs                         \\
tvs                      & 3650                     & trombosi venosa superficiale & superficial venous thrombosis \\
sn                       & 3055                     & sinistra                     & left                          \\
x                        & 2210                     & per                          & for                           \\
k                        & 2170                     & cancro                       & cancer                        \\
ivc                      & 2124                     & insufficienza venosa cronica & chronic venous insufficiency  \\
ii                       & 1768                     & inferiori                    & inferior                      \\
tep                      & 1695                     & tromboembolia polmonare      & pulmonary thromboembolism     \\
nao                      & 1291                     & nuovi anticoagulanti orali   & new oral anticoagulants       \\
ca                       & 1262                     & cancro                       & cancer                        \\
ds                       & 1001                     & destra                       & right                         \\
paz                      & 959                      & paziente                     & patient                       \\
tao                      & 924                      & terapia anticoagulante orale & oral anticoagulant therapy    \\
vgs                      & 911                      & vena grande safena           & great saphenous vein          \\
tv                       & 875                      & trombosi venosa              & venous thrombosis             \\
tp                       & 852                      & trombosi profonda            & deep thrombosis               \\
s                        & 819                      & sindrome                     & syndrome                      \\
dm                       & 800                      & diabete mellito              & diabetes mellitus             \\
bil                      & 792                      & bilaterale                   & bilateral                     \\
irc                      & 650                      & insufficienza renale cronica & chronic renal failure         \\
ins                      & 620                      & insufficienza                & failure                       \\
sup                      & 608                      & superiore                    & upper                         \\
fa                       & 592                      & fibrillazione atriale        & atrial fibrillation           \\
v                        & 576                      & venosa                       & venous                        \\
iv                       & 497                      & insufficienza venosa         & venous insufficiency          \\
spt                      & 469                      & sindrome post trombotica     & post-thrombotic syndrome      \\
npl                      & 451                      & neoplasia                    & neoplasm                      \\
iii                      & 444                      & inferiori                    & inferior                      \\
ep                       & 421                      & episodio                     & episode                       \\
tx                       & 364                      & trapianto                    & transplantation               \\
ctr                      & 351                      & controllo                    & control                       \\
art                      & 327                      & tromboembolismo venoso       & venous thromboembolism        \\
tpv                      & 326                      & trombosi profonda venosa     & deep vein thrombosis          \\
vs                       & 307                      & vena safena                  & saphenous vein                \\
cr                       & 297                      & cronica                      & chronic                       \\
gg                       & 269                      & giorni                       & days                          \\
tev                      & 256                      & tromboembolia venosa         & venous thromboembolism        \\
pta & 251 & angioplastica percutanea transluminale & percutaneous transluminal angioplasty \\
ven                      & 235                      & venosa                       & venous                        \\
int                      & 227                      & intervento                   & surgery                       \\
mvc                      & 225                      & malattia venosa cronica      & chronic venous disease        \\
ev                       & 220                      & eventuale                    & possible                      \\
tt                       & 203                      & tibio-tarsica                & tibio-tarsal                  \\ \bottomrule
\end{tabular}
\end{table}

\begin{table}[h!]
\centering
\caption{Sample of 50 misspelt words with the corresponding replacements applied by our pre-processing.}
\label{tab:typos}
\footnotesize
\begin{tabular}{lll}
\toprule
Original Word & Corrected Word & Corrected Word {[}ENG{]} \\
\midrule
iprtensioone           & ipertensione            & hypertension                      \\
stwnosi                & stenosi                 & stenosis                          \\
areomasica             & ateromasia              & atheromasia                       \\
aduta                  & caduta                  & fall                              \\
clacifiche             & calcifiche              & calcifications                    \\
lipitimiche            & lipotimie               & lipothymias                       \\
esignificative         & significative           & significant                       \\
veriginosa             & vertiginosa             & vertiginous                       \\
lipotimoco             & lipotimici              & lipotimics                        \\
emicrani               & emicrania               & migraine                          \\
cardiopatiia           & cardiopatia             & heart disease                     \\
30anni                 & anni                    & years                             \\
sinndrome              & sindrome                & syndrome                          \\
vertigicnosa           & vertiginosa             & vertiginous              \\
eriosa                 & arteriosa               & arterial                          \\
srteriopatia           & arteriopatia            & arteriopathy                      \\
carotgidee             & carotidea               & carotid                           \\
nvertiginosa           & vertiginosa             & vertiginous                       \\
nlipotimia             & lipotimici              & lipothymic                        \\
pregressotia           & pregresso               & previous                          \\
priomo                 & primo                   & first                             \\
dislipemua             & dislipidemia            & dyslipidaemia                     \\
caroltidea             & carotidea               & carotid                           \\
iportesa               & ipertesa                & hypertensive                      \\
cerebricolare          & cerebrovascolare        & cerebrovascular                   \\
neurologoin            & neurologo               & neurologist                       \\
cronarico              & coronarica              & coronary                          \\
arteiopatia            & arteriopatia            & arteriopathy                      \\
bilaterati             & bilaterale              & bilateral                         \\
aritimica              & aritmica                & arrhythmic                        \\
bpcoe                  & bpco                    & bpco                              \\
extrasisoliche         & extrasistolia           & extrasystole                      \\
paiente                & paziente                & patient                           \\
iprtnsione             & ipertensione            & hypertension                      \\
teriopati              & arteriopatia            & arteriopathy                      \\
caprtidea              & carotidea               & carotid                           \\
soszpetta              & sospetta                & suspected                         \\
dislidemica            & dislipidemia            & dyslipidaemia                     \\
lipotimicio            & lipotimici              & lipothymics                       \\
epeisodi               & episodi                 & episodes                          \\
ipertrigliceridiemia   & ipertrigliceridemia     & hypertriglyceridaemia             \\
ertiginosa             & vertiginosa             & vertigo                           \\
iperpensione           & ipertensione            & hypertension                      \\
neopla                 & neoplasia               & neoplasia                         \\
oderata                & moderata                & moderate                          \\
cardiopataia           & cardiopatia             & cardiopathy                       \\
polidistrettuae        & polidistrettuale        & polydistrict                      \\
dislipidemiaecocolo    & dislipidemico           & dyslipidaemic                     \\
ipwrtensone            & ipertensione            & hypertension                      \\
acugene                & acufene                 & tinnitus     \\ 
\bottomrule
\end{tabular}
\end{table}

\begin{table}[h!]
\centering
\caption{Results of clusters mapped to the disease labels on the manually labelled test sets, computed for clinical question (CQ) and referrals (REF) without applying pre-processing steps. P=Precision, R=Recall, F1=F1-Score. The values of all metrics are expressed as percentages.}
\label{tab:res-wo-preproc}
\begin{tabular}{lrrrrrr}
\toprule
                     & P-CQ           & R-CQ           & F1-CQ          & P-REF          & R-REF          & F1-REF         \\ \midrule
\multicolumn{7}{c}{\textit{ECD}}                                                                                           \\ \midrule
Umberto-E3C-Pipeline & \textbf{84.01} & \textbf{73.51} & \textbf{78.40} & \textbf{89.15} & \textbf{79.28} & \textbf{83.95} \\
Umberto-Base         & 77.99          & 70.68          & 74.18          & 85.01          & 73.18          & 78.66          \\
Word2Vec             & 75.63          & 62.25          & 68.27          & 78.90          & 67.77          & 72.99          \\
TF-IDF               & 46.28          & 33.61          & 38.92          & 45.73          & 37.83          & 41.46          \\
LDA                  & 24.38          & 32.27          & 27.88          & 16.51          & 25.97          & 20.24          \\ \midrule
\multicolumn{7}{c}{\textit{FEC}}                                                                                           \\ \midrule
Umberto-E3C-Pipeline & \textbf{88.59} & \textbf{85.53} & \textbf{87.12} & \textbf{90.99} & \textbf{87.65} & \textbf{89.32} \\
Umberto-Base         & 80.28          & 70.31          & 75.02          & 85.79          & 75.42          & 80.35          \\
Word2Vec             & 78.66          & 68.55          & 73.27          & 80.15          & 70.38          & 74.96          \\
TF-IDF               & 55.45          & 44.91          & 49.69          & 55.95          & 49.60          & 52.65          \\
LDA                  & 31.52          & 36.30          & 33.80          & 26.70          & 33.91          & 29.86          \\ \bottomrule
\end{tabular}
\end{table}

\clearpage

\section{Clustering}
\label{app:clustering}
\setcounter{figure}{0}  
\setcounter{table}{0}
\subsection{Hyperparameters and algorithms selection}
Table \ref{tab:hdbscan_hyper} reports the grid search for HDBSCAN hyperparameters, evaluated with F1-CQ on ECD dataset. 

\begin{longtable}[ptb]{@{}rrrr@{}}
\caption{Grid search for HDBSCAN hyperparameters, with F1-CQ on ECD dataset}
\label{tab:hdbscan_hyper} \\
   \toprule
\multicolumn{1}{l}{{min\_cluster\_size}} & \multicolumn{1}{l}{{min\_samples}} & \multicolumn{1}{l}{{cluster\_selection\_epsilon}} & {F1-CQ {[}\%{]}} \\ \midrule \endfirsthead 
   \toprule
\multicolumn{1}{l}{{min\_cluster\_size}} & \multicolumn{1}{l}{{min\_samples}} & \multicolumn{1}{l}{{cluster\_selection\_epsilon}} & {F1-CQ {[}\%{]}} \\ \midrule \endhead
    \\ \caption*{(continues)}\\
    \endfoot
    \\
    \endlastfoot
20                                              &                                           & 0                                                        & 11.42                   \\
20                                              &                                           & 0.1                                                      & 11.42                   \\
20                                              &                                           & 0.5                                                      & 11.42                   \\
20                                              & 1                                         & 0                                                        & 63.91                   \\
20                                              & 1                                         & 0.1                                                      & 63.91                   \\
20                                              & 1                                         & 0.5                                                      & 63.91                   \\
20                                              & 5                                         & 0                                                        & 10.13                   \\
20                                              & 5                                         & 0.1                                                      & 10.13                   \\
20                                              & 5                                         & 0.5                                                      & 10.13                   \\
20                                              & 10                                        & 0                                                        & 77.85                   \\
20                                              & 10                                        & 0.1                                                      & 77.85                   \\
20                                              & 10                                        & 0.5                                                      & 77.85                   \\
20                                              & 20                                        & 0                                                        & 11.42                   \\
20                                              & 20                                        & 0.1                                                      & 11.42                   \\
20                                              & 20                                        & 0.5                                                      & 11.42                   \\
50                                              &                                           & 0                                                        & 40.31                   \\
50                                              &                                           & 0.1                                                      & 40.31                   \\
50                                              &                                           & 0.5                                                      & 40.31                   \\
50                                              & 1                                         & 0                                                        & 60.81                   \\
50                                              & 1                                         & 0.1                                                      & 60.81                   \\
50                                              & 1                                         & 0.5                                                      & 60.81                   \\
50                                              & 5                                         & 0                                                        & 71.72                   \\
50                                              & 5                                         & 0.1                                                      & 71.72                   \\
50                                              & 5                                         & 0.5                                                      & 71.72                   \\
50                                              & 10                                        & 0                                                        & 70.40                   \\
50                                              & 10                                        & 0.1                                                      & 70.40                   \\
50                                              & 10                                        & 0.5                                                      & 70.40                   \\
50                                              & 20                                        & 0                                                        & 40.31                   \\
50                                              & 20                                        & 0.1                                                      & 40.31                   \\
50                                              & 20                                        & 0.5                                                      & 40.31                   \\
100                                             &                                           & 0                                                        & 23.87                   \\
100                                             &                                           & 0.1                                                      & 23.87                   \\
100                                             &                                           & 0.5                                                      & 23.87                   \\
100                                             & 1                                         & 0                                                        & 75.64                   \\
100                                             & 1                                         & 0.1                                                      & 75.64                   \\
100                                             & 1                                         & 0.5                                                      & 75.64                   \\
100                                             & 5                                         & 0                                                        & 44.60                   \\
100                                             & 5                                         & 0.1                                                      & 44.60                   \\
100                                             & 5                                         & 0.5                                                      & 44.60                   \\
100                                             & 10                                        & 0                                                        & 51.82                   \\
100                                             & 10                                        & 0.1                                                      & 51.82                   \\
100                                             & 10                                        & 0.5                                                      & 51.82                   \\
100                                             & 20                                        & 0                                                        & 40.31                   \\
100                                             & 20                                        & 0.1                                                      & 40.31                   \\
100                                             & 20                                        & 0.5                                                      & 40.31                   \\
200                                             &                                           & 0                                                        & 14.46                   \\
200                                             &                                           & 0.1                                                      & 14.46                   \\
200                                             &                                           & 0.5                                                      & 14.46                   \\
200                                             & 1                                         & 0                                                        & 67.41                   \\
200                                             & 1                                         & 0.1                                                      & 67.41                   \\
200                                             & 1                                         & 0.5                                                      & 67.41                   \\
200                                             & 5                                         & 0                                                        & 36.43                   \\
200                                             & 5                                         & 0.1                                                      & 36.43                   \\
200                                             & 5                                         & 0.5                                                      & 36.43                   \\
200                                             & 10                                        & 0                                                        & 43.32                   \\
200                                             & 10                                        & 0.1                                                      & 43.32                   \\
200                                             & 10                                        & 0.5                                                      & 43.32                   \\
200                                             & 20                                        & 0                                                        & 31.80                   \\
200                                             & 20                                        & 0.1                                                      & 31.80                   \\
200                                             & 20                                        & 0.5                                                      & 31.80                   \\
500                                             &                                           & 0                                                        & 18.64                   \\
500                                             &                                           & 0.1                                                      & 18.64                   \\
500                                             &                                           & 0.5                                                      & 18.64                   \\
500                                             & 1                                         & 0                                                        & 15.48                   \\
500                                             & 1                                         & 0.1                                                      & 15.48                   \\
500                                             & 1                                         & 0.5                                                      & 15.48                   \\
500                                             & 5                                         & 0                                                        & 14.45                   \\
500                                             & 5                                         & 0.1                                                      & 14.45                   \\
500                                             & 5                                         & 0.5                                                      & 14.45                   \\
500                                             & 10                                        & 0                                                        & 23.37                   \\
500                                             & 10                                        & 0.1                                                      & 23.37                   \\
500                                             & 10                                        & 0.5                                                      & 23.37                   \\
500                                             & 20                                        & 0                                                        & 16.31                   \\
500                                             & 20                                        & 0.1                                                      & 16.31                   \\
500                                             & 20                                        & 0.5                                                      & 16.31                   \\
step from 500                                   &                                           & 0                                                        & 34.52                   \\
step from 500                                   &                                           & 0.1                                                      & 34.52                   \\
step from 500                                   &                                           & 0.5                                                      & 34.52                   \\
step from 500                                   & 1                                         & 0                                                        & 71.54                   \\
step from 500                                   & 1                                         & 0.1                                                      & 71.54                   \\
step from 500                                   & 1                                         & 0.5                                                      & 71.54                   \\
step from 500                                   & 5                                         & 0                                                        & \textbf{79.48}                   \\
step from 500                                   & 5                                         & 0.1                                                      & \textbf{79.48}                   \\
step from 500                                   & 5                                         & 0.5                                                      & \textbf{79.48}                   \\
step from 500                                   & 10                                        & 0                                                        & 63.28                   \\
step from 500                                   & 10                                        & 0.1                                                      & 63.28                   \\
step from 500                                   & 10                                        & 0.5                                                      & 63.28                   \\
step from 500                                   & 20                                        & 0                                                        & 41.95                   \\
step from 500                                   & 20                                        & 0.1                                                      & 41.95                   \\
step from 500                                   & 20                                        & 0.5                                                      & 41.95                   \\ \bottomrule
\end{longtable}

\begin{figure}[h!]
    \centering
    \includegraphics[width=.9\linewidth]{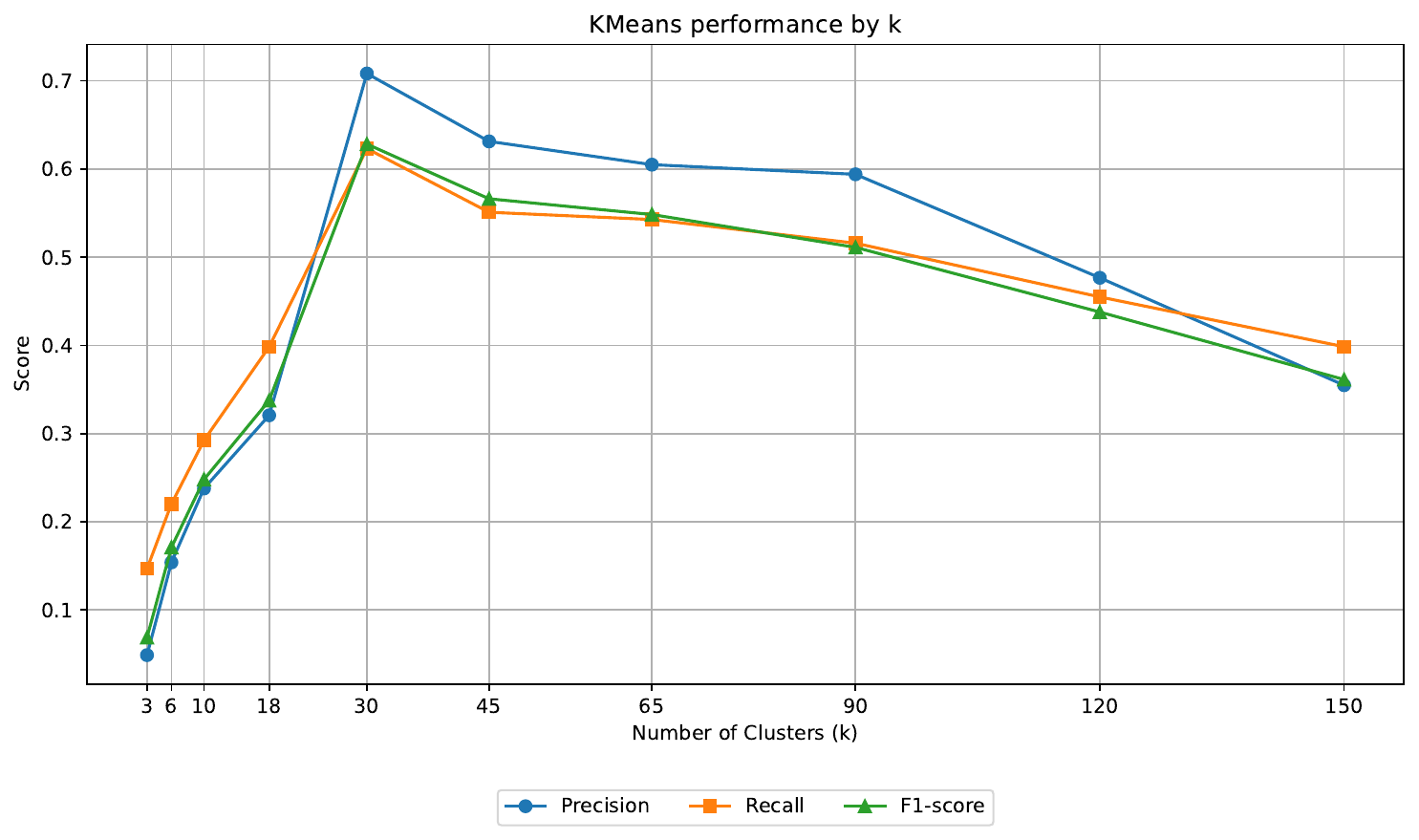}
    \caption{K-means performance on ECD-CQ varying the number of clusters}
    \label{fig:kmeans}
\end{figure}

\begin{figure}[h!]
    \centering
    \includegraphics[width=.9\linewidth]{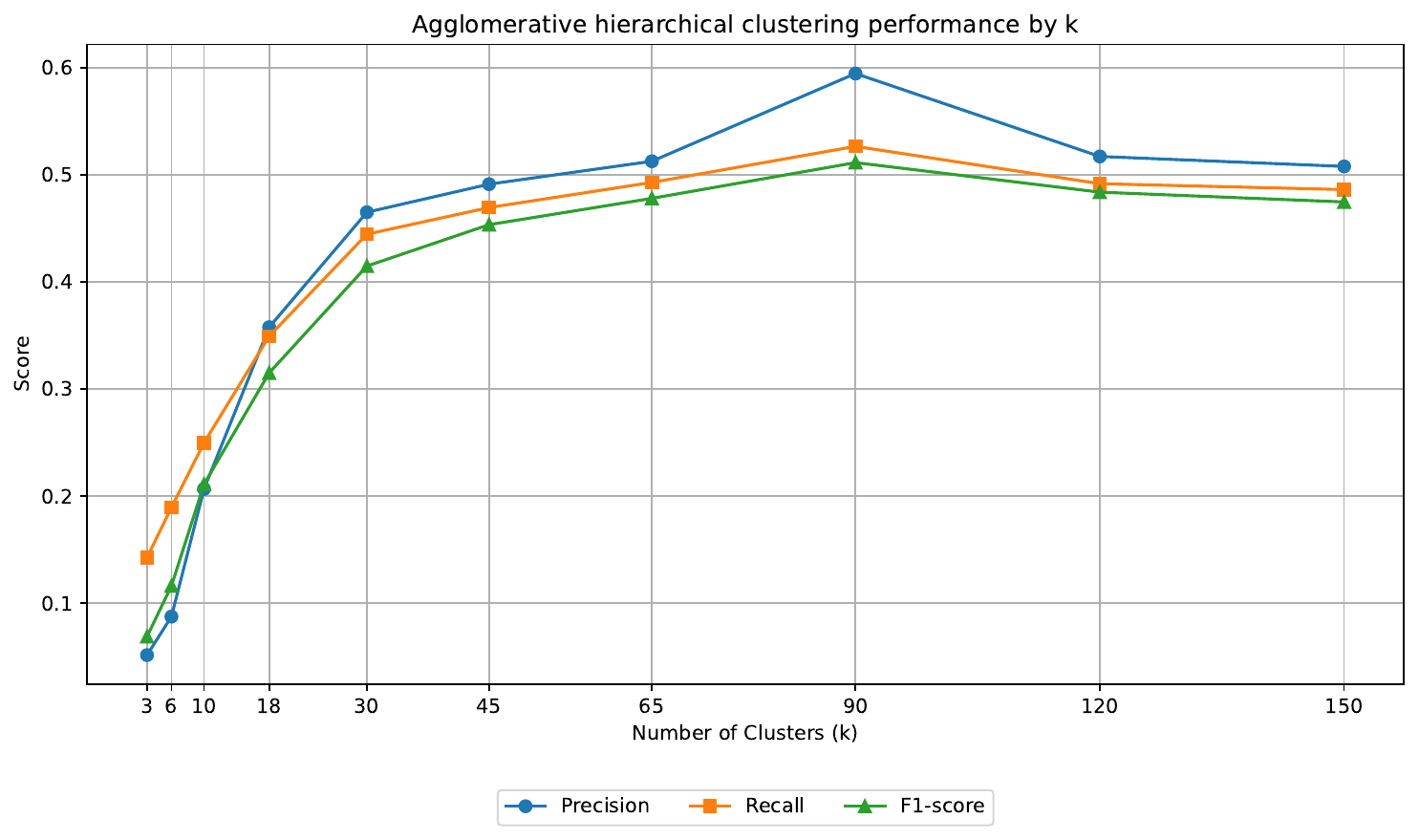}
    \caption{Agglomerative hierarchical clustering performance on ECD-CQ varying the number of clusters}
    \label{fig:hierarchical}
\end{figure}

Figures \ref{fig:kmeans} and \ref{fig:hierarchical} report the performance of K-Means and Agglomerative Hierarchical Clustering on the ECD dataset, at the CQ level, varying the number of clusters. Their results are consistently lower than those from HDBSCAN. 

For TF-IDF representation, we limited the vocabulary to the top 1,000 most frequent tokens. The best results presented are obtained with K-Means. The number of clusters was selected with the knee-elbow method analyzing WSS with k from 2 to 50, which led to k=40 for ECD and k=35 for FEC. 

For LDA, the number of topics (clusters) was selected by analyzing coherence with k from 2 to 50, which led to k=7 for ECD and k=9 for FEC.

For direct embeddings match, we matched Umberto embeddings of clinical questions with embeddings of known diseases/symptoms from guidelines, using cosine similarity. If no element in the guidelines matched with a similarity of at least 0.6 we assigned the \textit{Other} label.

\clearpage

\subsection{Additional clustering results}
Tables~\ref{tab:clusters-summ} and \ref{tab:clusters-summ-fec} present all the clusters identified for ECD and FEC, respectively, with their keyword descriptions, the sizes and the labels they were assigned.

\footnotesize
\begin{longtable}[hp]{@{}rp{3.5cm}p{3.5cm}p{2.5cm}rr@{}}
   \caption{Summary of all clusters identified for ECD, including their keyword representation, the label assigned them and the number of distinct clinical questions (CQ) and referrals (REF) that they contain.}
    \label{tab:clusters-summ} \\
   \toprule
ID & CLUSTER KEYWORDS {[}ITA{]} &
  CLUSTER KEYWORDS {[}ENG{]} &
  CLUSTER LABEL &
  \multicolumn{1}{l}{\# CQ} &
  \multicolumn{1}{l}{\# REF} \\ \midrule \endfirsthead 
   \toprule
ID & CLUSTER KEYWORDS {[}ITA{]} &
  CLUSTER KEYWORDS {[}ENG{]} &
  CLUSTER LABEL &
  \multicolumn{1}{l}{\# CQ} &
  \multicolumn{1}{l}{\# REF} \\ \midrule \endhead
    \\ \caption*{(continues)}\\
    \endfoot
    \\
    \endlastfoot
    \\ 1 & 
claudicatio; riferita &
  claudication; referred &
  claudication &
  1132 &
  1686 \\ 2 &
claudicatio; intermittens &
  claudicatio; intermittens &
  claudication &
  653 &
  1179 \\ 3 &
claudicatio &
  claudicatio &
  claudication &
  402 &
  530 \\ 4 &
claudicatio; sospetta; claudicazio &
  claudicatio; suspected; claudication &
  claudication &
  60 &
  126 \\ 5 &
claudicatio; arti; inferiori &
  claudicatio; limbs; lower &
  claudication &
  41 &
  49 \\ 6 &
dermatite &
  dermatitis &
  dermatitis &
  421 &
  474 \\ 7 &
diabete &
  diabetes &
  diabetes &
  1540 &
  3249 \\ 8 &
diabete; mellito &
  diabetes; mellitus &
  diabetes &
  534 &
  1847 \\ 9 &
diabete; cardiopatia; mellito;  &
  diabetes; heart disease; mellitus &
  diabetes &
  755 &
  1254 \\ 10 &
diabetico; diabetica &
  diabetes; diabetic &
  diabetes &
  209 &
  165 \\ 11 &
dislipidemia; diabete &
  dyslipidaemia; diabetes &
  diabetes &
  36 &
  68 \\ 12 &
dialisi &
  dialysis &
  dialysis &
  110 &
  222 \\ 13 &
dimero &
  dimer &
  dimer &
  1343 &
  1601 \\ 14 &
dispnea &
  dyspnoea &
  dyspnoea &
  104 &
  171 \\ 15 &
dolore &
  pain &
  pain &
  6343 &
  12051 \\ 16 &
deambulazione; dolore; difficolta &
  walking;  pain;  difficulty &
  pain &
  1294 &
  3884 \\ 17 &
algia; artralgia; piedi &
  algia; arthralgia; foot &
  pain &
  1028 &
  2491 \\ 18 &
polpaccio; dolore &
  calf; pain &
  pain &
  707 &
  1753 \\ 19 &
algie; arti; gambe; polpacci; bilaterali; natura; determinare &
  pain; limbs; calves; bilateral; nature; determine &
  pain &
  160 &
  631 \\ 20 &
dolore; natura; determinare; algie &
  pain; nature; determine; pain &
  pain &
  53 &
  126 \\ 21 &
dolore; popliteo &
  pain; popliteal &
  pain &
  35 &
  117 \\ 22 &
algia; covid; gamba &
  pain; covid; leg &
  pain &
  30 &
  63 \\ 23 &
dolore; polpaccio; natura; determinare &
  pain; calf; nature; determine &
  pain &
  12 &
  10 \\ 24 &
crampi; notturni &
  cramps; nocturnal &
  pain &
  6 &
  5 \\ 25 &
edema &
  edema &
  edema &
  11686 &
  20723 \\ 26 &
edemi; declivi &
  edema; declivities &
  edema &
  9491 &
  16473 \\ 27 &
edema; arti; edemi &
  edema; limbs; edemas &
  edema &
  3712 &
  7867 \\ 28 &
arti; inferiori; edemi &
  limbs; lower; edemas &
  edema &
  528 &
  2486 \\ 29 &
edema; gamba; piede; polpaccio &
  edema; leg; foot; calf &
  edema &
  1046 &
  2467 \\ 30 &
edema; arti; inferiori &
  edema;  limbs;  lower &
  edema &
  899 &
  2350 \\ 31 &
edema; perimalleolare &
  edema; perimalleolar &
  edema &
  1039 &
  2214 \\ 32 &
edema; natura; determinare &
  edema; nature; determine &
  edema &
  1027 &
  2199 \\ 33 &
edemi; perimalleolari &
  edema; perimalleolar &
  edema &
  1190 &
  2151 \\ 34 &
edema; malleolare &
  edema; malleolar &
  edema &
  935 &
  1136 \\ 35 &
edemi; vasculopatia; declivi; edema &
  edema; vasculopathy; declivities &
  edema &
  529 &
  602 \\ 36 &
edema; flebite &
  edema; phlebitis &
  edema &
  217 &
  297 \\ 37 &
arti; bilaterale; piede; edema; succulenza; perimalleolare &
  limbs; bilateral; foot; edema; succulence; perimalleolar &
  edema &
  117 &
  192 \\ 38 &
arti; inferiori; edema &
  limbs; lower; edema &
  edema &
  80 &
  168 \\ 39 &
arti; inferiori; natura; determinare; edema; edemi &
  limbs; lower; nature; determine; edema &
  edema &
  80 &
  168 \\ 40 &
edema; inferiori; arti &
  edema;  lower; limbs &
  edema &
  74 &
  163 \\ 41 &
edema; sinistra &
  edema; left &
  edema &
  61 &
  108 \\ 42 &
claudicatio; edemi; edema; cardiopatia; arti; inferiore; lieve &
  claudication; edema; edema; cardiopathy; limbs; lower; mild &
  edema &
  32 &
  34 \\ 43 &
linfedema; arti &
  lymphedema; limbs &
  edema &
  23 &
  24 \\ 44 &
emorragia &
  haemorrhage &
  haemorrhage &
  18 &
  29 \\ 45 &
formicolio &
  tingling &
  tingling &
  43 &
  170 \\ 46 &
freddo &
  cold &
  cold &
  49 &
  147 \\ 47 &
fumatore &
  smoker &
  smoker &
  121 &
  193 \\ 48 &
gravidanza &
  pregnancy &
  pregnancy &
  92 &
  225 \\ 49 &
insufficienza respiratoria &
  respiratory insufficiency &
  respiratory insufficiency &
  724 &
  1060 \\ 50 &
venosa; insufficienza &
  venous; insufficiency &
  venous insufficiency &
  30324 &
  71150 \\ 51 &
venosa; arti; inferiori; insufficienza; sospetta &
  venous; limbs; lower; insufficiency; suspected &
  venous insufficiency &
  19532 &
  28831 \\ 52 &
superficiale; venosa; insufficienza &
  superficial; venous; insufficiency &
  venous insufficiency &
  276 &
  613 \\ 53 &
insufficienza; venosa; cronica &
  venous; insufficiency; chronic &
  venous insufficiency &
  415 &
  506 \\ 54 &
venosa; safenectomia; insufficienza &
  venous; saphenectomy; insufficiency &
  venous insufficiency &
  280 &
  311 \\ 55 &
venosa; claudicatio; insufficienza &
  venous; claudication; insufficiency &
  venous insufficiency &
  109 &
  180 \\ 56 &
venosa; arti; insufficienza; varici &
  venous; limb; insufficiency; varicose &
  venous insufficiency &
  110 &
  112 \\ 57 &
insufficienza; venosa; arti; inferiori &
  venous; insufficiency; limbs; lower &
  venous insufficiency &
  23 &
  102 \\ 58 &
venoso; circolo; deficit; arti;  &
  venous; circulation; insufficiency; limbs &
  venous insufficiency &
  74 &
  102 \\ 59 &
venosa; insufficienza; arti; inferiori; edemi &
  venous; insufficiency; limbs; lower; edemas &
  venous insufficiency &
  47 &
  58 \\ 60 &
insufficienza; venosa &
  venous; insufficiency &
  venous insufficiency &
  0 &
  0 \\ 61 &
arti; venosa; inferiori &
  limbs; venous; lower &
  venous insufficiency &
  0 &
  0 \\ 62 &
venosa; edemi; declivi; insufficienza &
  venous; edemas; declivity; insufficiency &
  venous insufficiency &
  0 &
  0 \\ 63 &
venosa; insufficienza; edemi; arti; inferiori &
  venous; insufficiency; edemas; limbs; lower &
  venous insufficiency &
  0 &
  0 \\ 64 &
venosa; cronica &
  venous; chronic &
  venous insufficiency &
  0 &
  0 \\ 65 &
insufficienza; venosa; cronica; sospetta &
  venous; insufficiency; chronic; suspected &
  venous insufficiency &
  264 &
  374 \\ 66 &
intervento; ginocchio; artroprotesi; valutazione &
  surgery; knee; arthroplasty; evaluation &
  intervention &
  67 &
  77 \\ 67 &
lista; trapianto; rene; inserimento &
  list; transplant; kidney; insertion &
  intervention &
  61 &
  69 \\ 68 &
lista; trapianto; valutazione; inserimento; rene; sinistra; dialisi; intervento &
  list; transplant; evaluation; insertion; kidney; left; dialysis; intervention &
  intervention &
  49 &
  52 \\ 69 &
ipercolesterolemia &
  hypercholesterolaemia &
  hypercholesterolaemia &
  363 &
  416 \\ 70 &
ipoestesia &
  hypoesthesia &
  hypoesthesia &
  31 &
  80 \\ 71 &
ischemia &
  ischaemia &
  ischaemia &
  517 &
  617 \\ 72 &
parestesie &
  paresthesias &
  paresthesias &
  1834 &
  3965 \\ 73 &
parestesie; gamba; sinistra &
  paresthesias; leg; left &
  paresthesias &
  541 &
  891 \\ 74 &
pesantezza &
  heaviness &
  heaviness &
  215 &
  475 \\ 75 &
arti; inferiori; linfedema; pesantezza &
  limbs; lower; lymphedema; heaviness &
  heaviness &
  25 &
  124 \\ 76 &
arti; inferiori; pesantezza &
  limbs; lower; heaviness &
  heaviness &
  18 &
  106 \\ 77 &
safenectomia &
  saphenectomy &
  saphenectomy &
  1988 &
  4177 \\ 78 &
safenectomia; esiti &
  saphenectomy; outcomes &
  saphenectomy &
  1374 &
  3617 \\ 79 &
safenectomia; pregressa &
  saphenectomy; previous &
  saphenectomy &
  342 &
  379 \\ 80 &
safenectomia; edemi &
  saphenectomy; edemas &
  saphenectomy &
  41 &
  117 \\ 81 &
sindrome metabolica &
  metabolic syndrome &
  metabolic syndrome &
  12 &
  48 \\ 82 &
stent &
  stent &
  stent &
  301 &
  612 \\ 83 &
tromboflebite &
  thrombophlebitis &
  thrombophlebitis &
  7073 &
  17415 \\ 84 &
flebite; sospetta &
  phlebitis; suspected &
  thrombophlebitis &
  5408 &
  12379 \\ 85 &
flebite; gamba; superficiale &
  phlebitis; leg; superficial &
  thrombophlebitis &
  3037 &
  8850 \\ 86 &
varicoflebite; sospetta; tromboflebite &
  varicophlebitis; suspected; thrombophlebitis &
  thrombophlebitis &
  4546 &
  7739 \\ 87 &
tromboflebite; superficiale &
  thrombophlebitis; superficial &
  thrombophlebitis &
  950 &
  1648 \\ 88 &
gamba; tromboflebite; destra &
  leg; thrombophlebitis; right &
  thrombophlebitis &
  397 &
  486 \\ 89 &
tromboflebite; safena &
  thrombophlebitis; saphenous vein &
  thrombophlebitis &
  386 &
  384 \\ 90 &
flebite; coscia; superficiale &
  phlebitis;  thigh;  superficial &
  thrombophlebitis &
  109 &
  194 \\ 91 &
flebopatia; sospetta; arti; inferiori; sospetta &
  phlebopathy; suspected; lower; limbs &
  thrombophlebitis &
  209 &
  175 \\ 92 &
tromboflebite; poplitea; arti; inferiori &
  thrombophlebitis; popliteal; lower; limbs &
  thrombophlebitis &
  54 &
  102 \\ 93 &
flebite; superficiale; gamba &
  phlebitis; superficial; leg &
  thrombophlebitis &
  48 &
  97 \\ 94 &
tromboflebite; sospetta;  &
  thrombophlebitis; suspected &
  thrombophlebitis &
  30 &
  68 \\ 95 &
trombosi &
  thrombosis &
  thrombosis &
  17060 &
  43558 \\ 96 &
trombosi; venosa; profonda &
  thrombosis; venous; deep &
  thrombosis &
  17453 &
  38230 \\ 97 &
trombosi; vascolare;  &
  thrombosis; vascular &
  thrombosis &
  5086 &
  10180 \\ 98 &
trombosi; venosa &
  venous; thrombosis &
  thrombosis &
  1325 &
  3337 \\ 99 &
venosa; trombosi; profonda; sospetto; segni; clinici; clinico &
  venous; thrombosis; deep; suspected; signs; clinical &
  thrombosis &
  566 &
  1152 \\ 100 &
poplitea; trombosi &
  popliteal; thrombosis &
  thrombosis &
  607 &
  642 \\ 101 &
venosa; trombosi; poplitea &
  venous; thrombosis; popliteal &
  thrombosis &
  407 &
  525 \\ 102 &
trombosi; safena; insufficienza &
  thrombosis; saphenous; insufficiency &
  thrombosis &
  362 &
  472 \\ 103 &
segni; clinici; sospetto; clinico; trombosi; venosa; profonda; edema &
  signs; clinical; suspected; clinical; venous; thrombosis; deep; edema &
  thrombosis &
  342 &
  389 \\ 104 &
safena; trombosi &
  saphenous vein; thrombosis &
  thrombosis &
  301 &
  263 \\ 105 &
post; trombotica &
  post; thrombotic &
  thrombosis &
  36 &
  73 \\ 106 &
tumefazione &
  swelling &
  swelling &
  533 &
  741 \\ 107 &
tumefazione; gamba &
  swelling; leg &
  swelling &
  172 &
  448 \\ 108 &
destro; polpaccio; tumefazione; paziente &
  right; calf; swelling; patient &
  swelling &
  35 &
  49 \\ 109 &
ulcera; vascolare &
  ulcer; vascular &
  ulcer &
  828 &
  3861 \\ 110 &
ulcera &
  ulcer &
  ulcer &
  1258 &
  2411 \\ 111 &
varici &
  varices &
  varices &
  13999 &
  35939 \\ 112 &
arti; inferiori; varici &
  limbs; lower; varices &
  varices &
  7018 &
  18157 \\ 113 &
venose; varici &
  venous; varices &
  varices &
  4957 &
  13554 \\ 114 &
varicosita; stasi &
  varicosities; stasis &
  varices &
  957 &
  2701 \\ 115 &
varici; arti; inferiori &
  varices; limbs; lower limbs &
  varices &
  219 &
  277 \\ 116 &
vene; varicose; algie &
  veins; varicose; algie &
  varices &
  121 &
  194 \\ 117 &
safena; varici &
  saphenous veins; varicose &
  varices &
  47 &
  136 \\ 118 &
varicosita; varicosita' &
  varicosities; varicositis &
  varices &
  86 &
  97 \\ 119 &
vascolari; esiti; polmonite; varici; chirurgici; linfangite &
  vascular; outcomes; pneumonia; varices; surgical; lymphangitis &
  varices &
  60 &
  88 \\ 120 &
varici; sospette &
  varicose veins; suspected &
  varices &
  34 &
  63 \\ 121 &
varicosita; gamba &
  varicosities; leg &
  varices &
  80 &
  63 \\ 122 &
varici; intervento &
  varicosities; surgery &
  varices intervention &
  3144 &
  7126 \\ 123 &
vasculopatia; periferica &
  vasculopathy; peripheral &
  vasculopathy &
  3820 &
  8200 \\ 124 &
vasculopatia &
  vasculopathy; peripheral &
  vasculopathy &
  769 &
  1886 \\ 125 &
vasculopatia; diabetico &
  vasculopathy; diabetic &
  vasculopathy &
  393 &
  408 \\ 126 &
arteriopatia; obliterante &
  arteriopathy; obliterative &
  vasculopathy &
  288 &
  272 \\ 127 &
arteriopatia; obliterante; periferica &
  arteriopathy; obliterative; peripheral &
  vasculopathy &
  47 &
  117 \\ 128 &
iperteso; vasculopatico; vasculopatia; dislipidemico &
  hypertensive; vasculopathy; dyslipidemic &
  vasculopathy &
  17 &
  97 \\ 129 &
aneurisma; aorta; esiti; popliteo; addominale; aortico &
  aneurysm; aorta; popliteal; abdominal; aortic &
  vasculopathy &
  42 &
  78 \\ 130 &
vasculopatia; arti; inferiori &
  vasculopathy; limbs; lower &
  vasculopathy &
  36 &
  78 \\ 131 &
vasculopatia; arti; inferiori; polidistrettuale &
  vasculopathy; limb; lower; polydistrict &
  vasculopathy &
  35 &
  49 \\ 132 &
vasculopatia; venosa &
  vasculopathy; venous &
  vasculopathy &
  25 &
  34 \\ 133 &
arteriopatia &
  arteriopathy &
  vasculopathy &
  20 &
  25 \\ 134 &
natura; determinare; periferica &
  nature; determine; peripheral &
  other &
  3978 &
  10039 \\ 135 &
natura; determinare; arti; inferiori &
  nature; determine; limbs; lower &
  other &
  5018 &
  9601 \\ 136 &
paziente; polsi; periferici &
  patient; wrists; peripheral &
  other &
  2018 &
  8015 \\ 137 &
arti; inferiori &
  limbs; lower &
  other &
  2190 &
  5165 \\ 138 &
safena; incontinenza &
  saphenous;  incontinence &
  other &
  288 &
  782 \\ 139 &
ipertrofia &
  hypertrophy &
  hypertrophy &
  506 &
  591 \\ 140 &
tromboembolia; polmonare; esiti &
  thromboembolism; pulmonary; outcomes &
  other &
  386 &
  379 \\ 141 &
embolia; polmonare &
  embolism; pulmonary &
  embolism &
  110 &
  174 \\ 142 &
femoro; femorale; paziente; popliteo; sospetta; asse; esiti; venosa &
  femoral; femoral; patient; popliteal; suspected; axis; outcomes; venous &
  other &
  49 &
  49 \\ 143 &
stenosi; venoso &
  stenosis; venous &
  other &
  30 &
  49 \\ 144 &
dislipidemia; ipertensione &
  dyslipidaemia; hypertension &
  dyslipidaemia &
  25 &
  44 \\ 145 &
aorta; sottorenale &
  aorta; subrenal &
  other &
  25 &
  29 \\ 
  \bottomrule
\end{longtable}

\footnotesize
\begin{longtable}[hp]{@{}r p{3.5cm} p{3.5cm} p{2.5cm} rr@{}}
   \caption{Summary of all clusters identified for FEC, including their keyword representation, the label assigned to them and the number of distinct clinical questions (CQ) and referrals (REF) that they contain.}
    \label{tab:clusters-summ-fec}\\
  \toprule
ID & CLUSTER KEYWORDS {[}ITA{]} & CLUSTER KEYWORDS {[}ENG{]} & CLUSTER LABEL & \# CQ & \# REF \\ \midrule 1 &
addominalgia; alvo &
  abdominalgia; alvo &
  abdominal pain &
  \multicolumn{1}{r}{163} &
  \multicolumn{1}{r}{371} \\ 2 & 
addominalgie; natura; determinare &
  abdominalgia; nature; determine &
  abdominal pain &
  \multicolumn{1}{r}{714} &
  \multicolumn{1}{r}{5.801} \\ 3 &
addominali; algie; dolori &
  abdominalgia; algia; pain &
  abdominal pain &
  \multicolumn{1}{r}{894} &
  \multicolumn{1}{r}{5.430} \\ 4 &
addominali; coliche &
  abdominal; colic &
  abdominal pain &
  \multicolumn{1}{r}{240} &
  \multicolumn{1}{r}{3.775} \\ 5 &
alvo; addominalgie &
  bowel; abdominal pain &
  abdominal pain &
  \multicolumn{1}{r}{224} &
  \multicolumn{1}{r}{2.507} \\ 6 &
calo; ponderale; addominalgia &
  fall; weight; abdominalgia &
  abdominal pain &
  \multicolumn{1}{r}{100} &
  \multicolumn{1}{r}{215} \\ 7 &
natura; determinare; coliche &
  nature; determine; colic &
  abdominal pain &
  \multicolumn{1}{r}{325} &
  \multicolumn{1}{r}{2.590} \\ 8 &
natura; determinare; addominali; coliche &
  nature; determine; abdominal; colic &
  abdominal pain &
  \multicolumn{1}{r}{3.687} &
  \multicolumn{1}{r}{5.605} \\ 9 &
fianco; dolore; persistente &
  flank; pain; persistent &
  abdominal pain &
  \multicolumn{1}{r}{61} &
  \multicolumn{1}{r}{121} \\ 10 &
fossa; addominalgia &
  fossa; abdominalgia &
  abdominal pain &
  \multicolumn{1}{r}{192} &
  \multicolumn{1}{r}{2.125} \\ 11 &
fossa; iliaca; algia &
  fossa; iliac; algia &
  abdominal pain &
  \multicolumn{1}{r}{90} &
  \multicolumn{1}{r}{194} \\ 12 &
fossa; iliaca; algie &
  fossa; iliac; algia &
  abdominal pain &
  \multicolumn{1}{r}{33} &
  \multicolumn{1}{r}{71} \\ 13 &
fossa; iliaca; dolore &
  fossa; iliac; pain &
  abdominal pain &
  \multicolumn{1}{r}{302} &
  \multicolumn{1}{r}{662} \\ 14 &
ipocondrio; destra; dolore; sinistra &
  hypochondrium; right; pain; left &
  abdominal pain &
  \multicolumn{1}{r}{127} &
  \multicolumn{1}{r}{281} \\ 15 &
quadranti; addominalgia; addominalgie &
  quadrant; abdominalgia; abdominalgia &
  abdominal pain &
  \multicolumn{1}{r}{33} &
  \multicolumn{1}{r}{74} \\ 16 &
stipsi; addominalgia &
  constipation; abdominalgia &
  abdominal pain &
  \multicolumn{1}{r}{501} &
  \multicolumn{1}{r}{1.105} \\ 17 &
stipsi; addominali &
  constipation; abdominal &
  abdominal pain &
  \multicolumn{1}{r}{112} &
  \multicolumn{1}{r}{250} \\ 18 &
stipsi; addominali; algie &
  constipation; abdominal; pain &
  abdominal pain &
  \multicolumn{1}{r}{61} &
  \multicolumn{1}{r}{139} \\ 19 &
alvo; alterazione; alterazioni &
  alvo; alteration &
  altered bowel &
  \multicolumn{1}{r}{432} &
  \multicolumn{1}{r}{3.500} \\ 20 &
alvo; alterno &
  bowel; abdominalgia &
  altered bowel &
  \multicolumn{1}{r}{610} &
  \multicolumn{1}{r}{2.496} \\ 21 &
alvo; alterno; addominalgia &
  bowel; abdominalgia &
  altered bowel &
  \multicolumn{1}{r}{287} &
  \multicolumn{1}{r}{1.925} \\ 22 &
alvo; calo; ponderale &
  bowel; decline; weight &
  altered bowel &
  \multicolumn{1}{r}{196} &
  \multicolumn{1}{r}{1.655} \\ 23 &
alvo; stipsi &
  bowel; constipation &
  altered bowel &
  \multicolumn{1}{r}{129} &
  \multicolumn{1}{r}{277} \\ 24 &
fossa; alvo &
  fossa; alvo &
  altered bowel &
  \multicolumn{1}{r}{197} &
  \multicolumn{1}{r}{1.080} \\ 25 &
anemia; emoglobina &
  anaemia; haemoglobin &
  anaemia &
  \multicolumn{1}{r}{61} &
  \multicolumn{1}{r}{128} \\ 26 &
anemia; natura; determinare; grave &
  anaemia; nature; determine; severe &
  anaemia &
  \multicolumn{1}{r}{602} &
  \multicolumn{1}{r}{4.392} \\ 27 &
anemia; sangue; occulto; positivo &
  anaemia; blood; occult; positive &
  anaemia &
  \multicolumn{1}{r}{315} &
  \multicolumn{1}{r}{2.366} \\ 28 &
anemia; sideropenica; origine; indeterminata &
  anaemia; sideropenic; origin; undetermined &
  anaemia &
  \multicolumn{1}{r}{332} &
  \multicolumn{1}{r}{6.663} \\ 29 &
anemia; sideropenica; sangue; occulto &
  anaemia; sideropenic; blood; occult &
  anaemia &
  \multicolumn{1}{r}{49} &
  \multicolumn{1}{r}{112} \\ 30 &
anemizzazione; grave; severa; sangue &
  anaemia; severe; blood &
  anaemia &
  \multicolumn{1}{r}{456} &
  \multicolumn{1}{r}{3.814} \\ 31 &
anemizzazione; natura; determinare &
  anaemia; nature; determine &
  anaemia &
  \multicolumn{1}{r}{250} &
  \multicolumn{1}{r}{2.504} \\ 32 &
sideropenica; anemia &
  iron deficiency; anaemia &
  anaemia &
  \multicolumn{1}{r}{365} &
  \multicolumn{1}{r}{3.154} \\ 33 &
diarrea; colite; cronica &
  diarrhoea; colitis; chronic &
  chronic diarrhoea &
  \multicolumn{1}{r}{95} &
  \multicolumn{1}{r}{757} \\ 34 &
diarrea; cronica &
  diarrhoea; chronic &
  chronic diarrhoea &
  \multicolumn{1}{r}{778} &
  \multicolumn{1}{r}{7.668} \\ 35 &
diarroico; alvo &
  diarrhoeic; alvo &
  chronic diarrhoea &
  \multicolumn{1}{r}{193} &
  \multicolumn{1}{r}{2.694} \\ 36 &
natura; determinare; diarrea; dolore; addominale &
  nature; determine; diarrhoea; pain; abdominal &
  chronic diarrhoea &
  \multicolumn{1}{r}{409} &
  \multicolumn{1}{r}{2.414} \\ 37 &
colite; ischemica &
  colitis; ischaemic &
  chronic inflammatory bowel disease &
  \multicolumn{1}{r}{119} &
  \multicolumn{1}{r}{286} \\ 38 &
crohn; morbo &
  crohn disease; disease &
  chronic inflammatory bowel disease &
  \multicolumn{1}{r}{540} &
  \multicolumn{1}{r}{3.812} \\ 39 &
crohn; morbo; ileo &
  crohn; disease; ileum &
  chronic inflammatory bowel disease &
  \multicolumn{1}{r}{71} &
  \multicolumn{1}{r}{202} \\ 40 &
infiammatoria; malattia; intestinale; cronica &
  inflammatory; disease; intestinal; chronic &
  chronic inflammatory bowel disease &
  \multicolumn{1}{r}{220} &
  \multicolumn{1}{r}{2.333} \\ 41 &
malattie; infiammatorie; croniche; intestinali &
  diseases; inflammatory; chronic; intestinal; disease &
  chronic inflammatory bowel disease &
  \multicolumn{1}{r}{313} &
  \multicolumn{1}{r}{3.780} \\ 42 &
malattie; infiammatorie; croniche; intestinali; calprotectina &
  inflammatory; disease; chronic; intestinal; calprotectin &
  chronic inflammatory bowel disease &
  \multicolumn{1}{r}{19} &
  \multicolumn{1}{r}{51} \\ 43 &
stipsi; colite &
  constipation; colitis &
  chronic inflammatory bowel disease &
  \multicolumn{1}{r}{74} &
  \multicolumn{1}{r}{185} \\ 44 &
ulcerosa; colite &
  ulcerative; colitis &
  chronic inflammatory bowel disease &
  \multicolumn{1}{r}{2.985} &
  \multicolumn{1}{r}{7.323} \\ 45 &
ulcerosa; rettocolite &
  ulcerative; rectocolitis &
  chronic inflammatory bowel disease &
  \multicolumn{1}{r}{283} &
  \multicolumn{1}{r}{3.498} \\ 46 &
adenocarcinoma; colon &
  adenocarcinoma; colon &
  colon cancer &
  \multicolumn{1}{r}{336} &
  \multicolumn{1}{r}{753} \\ 47 &
adenoma; adenomi; colon &
  adenoma; adenomas; colon &
  colon cancer &
  \multicolumn{1}{r}{50} &
  \multicolumn{1}{r}{102} \\ 48 &
alvo; tumore; colon &
  bowel; cancer; colon &
  colon cancer &
  \multicolumn{1}{r}{57} &
  \multicolumn{1}{r}{120} \\ 49 &
anamnesi; colon &
  anamnesis; colon &
  colon cancer &
  \multicolumn{1}{r}{76} &
  \multicolumn{1}{r}{301} \\ 50 &
colico; tumore &
  colic; cancer &
  colon cancer &
  \multicolumn{1}{r}{97} &
  \multicolumn{1}{r}{219} \\ 51 &
colon; adenoma; adenomi; linfoma &
  colon; adenoma; lymphoma &
  colon cancer &
  \multicolumn{1}{r}{123} &
  \multicolumn{1}{r}{394} \\ 52 &
colon; discendente; tumore &
  colon; descending; cancer &
  colon cancer &
  \multicolumn{1}{r}{61} &
  \multicolumn{1}{r}{137} \\ 53 &
colon; pregressa; asportazione; tumore &
  colon; previous; removal; cancer &
  colon cancer &
  \multicolumn{1}{r}{41} &
  \multicolumn{1}{r}{80} \\ 54 &
colon; ricerca; resezione; dopo; sincrone; cancro; lesioni; curativa &
  colon; research; resection; after; synchronous; cancer; lesions; curative &
  colon cancer &
  \multicolumn{1}{r}{171} &
  \multicolumn{1}{r}{2.878} \\ 55 &
ematochezia; tumore; colon &
  haematochezia; cancer; colon &
  colon cancer &
  \multicolumn{1}{r}{389} &
  \multicolumn{1}{r}{867} \\ 56 &
emicolectomia; tumore; esiti; destra; colon &
  haemolectomy; cancer; outcomes; right; colon &
  colon cancer &
  \multicolumn{1}{r}{452} &
  \multicolumn{1}{r}{1.055} \\ 57 &
neoformazione; rettale; retto; colon &
  neoformation; rectal; rectum; colon &
  colon cancer &
  \multicolumn{1}{r}{51} &
  \multicolumn{1}{r}{119} \\ 58 &
neoplasia; colon &
  neoplasia; colon &
  colon cancer &
  \multicolumn{1}{r}{438} &
  \multicolumn{1}{r}{1.323} \\ 59 &
neoplasia; eteroplasia &
  neoplasm; heteroplasm &
  colon cancer &
  \multicolumn{1}{r}{113} &
  \multicolumn{1}{r}{289} \\ 60 &
sigma; tumore &
  sigma; cancer &
  colon cancer &
  \multicolumn{1}{r}{197} &
  \multicolumn{1}{r}{449} \\ 61 &
stipsi; tumore; colon &
  constipation; cancer; colon &
  colon cancer &
  \multicolumn{1}{r}{286} &
  \multicolumn{1}{r}{682} \\ 62 &
tumore; colon &
  cancer; colon &
  colon cancer &
  \multicolumn{1}{r}{5.725} &
  \multicolumn{1}{r}{16.757} \\ 63 &
tumore; colon; controllo &
  cancer; colon; control &
  colon cancer &
  \multicolumn{1}{r}{108} &
  \multicolumn{1}{r}{324} \\ 64 &
tumore; colon; famigliarita &
  cancer; colon; familial &
  colon cancer &
  \multicolumn{1}{r}{346} &
  \multicolumn{1}{r}{1.111} \\ 65 &
tumore; intestinale &
  cancer; intestine &
  colon cancer &
  \multicolumn{1}{r}{305} &
  \multicolumn{1}{r}{840} \\ 66 &
tumore; pregresso; colon &
  cancer; proctorragia; colon &
  colon cancer &
  \multicolumn{1}{r}{306} &
  \multicolumn{1}{r}{815} \\ 67 &
tumore; proctorragia; colon &
  cancer; proctorragia; colon &
  colon cancer &
  \multicolumn{1}{r}{71} &
  \multicolumn{1}{r}{148} \\ 68 &
tumore; retto &
  cancer; rectum &
  colon cancer &
  \multicolumn{1}{r}{266} &
  \multicolumn{1}{r}{671} \\ 69 &
ostinata; stipsi &
  obstinate; constipation &
  constipation &
  \multicolumn{1}{r}{2.356} &
  \multicolumn{1}{r}{5.962} \\ 70 &
stipsi; cronica &
  constipation; chronic &
  constipation &
  \multicolumn{1}{r}{745} &
  \multicolumn{1}{r}{7.521} \\ 71 &
stipsi; fossa; iliaca &
  constipation; iliac fossa &
  constipation &
  \multicolumn{1}{r}{27} &
  \multicolumn{1}{r}{59} \\ 72 &
stipsi; ostinata; addominali &
  constipation; obstinate; abdominal &
  constipation &
  \multicolumn{1}{r}{88} &
  \multicolumn{1}{r}{190} \\ 73 &
stipsi; ponderale; calo &
  constipation; ponderal; fall &
  constipation &
  \multicolumn{1}{r}{122} &
  \multicolumn{1}{r}{1.377} \\ 74 &
diverticolite &
  diverticulitis &
  diverticulitis &
  \multicolumn{1}{r}{759} &
  \multicolumn{1}{r}{3.148} \\ 75 &
diverticolite; acuta; recente &
  diverticulitis; acute; recent &
  diverticulitis &
  \multicolumn{1}{r}{373} &
  \multicolumn{1}{r}{1.172} \\ 76 &
sigma; diverticolite &
  sigma; diverticulitis &
  diverticulitis &
  \multicolumn{1}{r}{101} &
  \multicolumn{1}{r}{221} \\ 77 &
sigmoidite; sospetta; recente; episodio; diverticolare; esiti; diverticolite; diverticolosi &
  sigmoiditis; suspected; recent; episode; diverticular; outcome; diverticulitis; diverticulosis &
  diverticulitis &
  \multicolumn{1}{r}{55} &
  \multicolumn{1}{r}{128} \\ 78 &
alvo; diverticolosi &
  bowel; diverticulosis &
  diverticulosis &
  \multicolumn{1}{r}{359} &
  \multicolumn{1}{r}{2.126} \\ 79 &
diverticolare; malattia &
  diverticular; disease &
  diverticulosis &
  \multicolumn{1}{r}{147} &
  \multicolumn{1}{r}{438} \\ 80 &
diverticoli; diverticolare; colon; poliposi &
  diverticula; diverticular; colon; polyposis &
  diverticulosis &
  \multicolumn{1}{r}{388} &
  \multicolumn{1}{r}{1.849} \\ 81 &
diverticolosi; addominalgia &
  diverticulosis; abdominalgia &
  diverticulosis &
  \multicolumn{1}{r}{355} &
  \multicolumn{1}{r}{782} \\ 82 &
diverticolosi; addominali &
  diverticulosis; abdominal &
  diverticulosis &
  \multicolumn{1}{r}{264} &
  \multicolumn{1}{r}{2.213} \\ 83 &
diverticolosi; colon &
  diverticulosis; colon &
  diverticulosis &
  \multicolumn{1}{r}{337} &
  \multicolumn{1}{r}{3.281} \\ 84 &
diverticolosi; colon; diverticolite &
  diverticulosis; colon; diverticulitis &
  diverticulosis &
  \multicolumn{1}{r}{253} &
  \multicolumn{1}{r}{754} \\ 85 &
diverticolosi; sigma &
  diverticulosis; sigma &
  diverticulosis &
  \multicolumn{1}{r}{117} &
  \multicolumn{1}{r}{256} \\ 86 &
pregressa; asportazione; diverticolosi &
  previous; removal; diverticulosis &
  diverticulosis &
  \multicolumn{1}{r}{412} &
  \multicolumn{1}{r}{2.096} \\ 87 &
pregressa; polipectomia; diverticolosi &
  previous; polypectomy; diverticulosis &
  diverticulosis &
  \multicolumn{1}{r}{142} &
  \multicolumn{1}{r}{306} \\ 88 &
stipsi; diverticolosi &
  constipation; diverticulosis &
  diverticulosis &
  \multicolumn{1}{r}{136} &
  \multicolumn{1}{r}{957} \\ 89 &
emorroidaria; sindrome; patologia &
  haemorrhoidal; syndrome; pathology &
  haemorrhoidal disease &
  \multicolumn{1}{r}{113} &
  \multicolumn{1}{r}{259} \\ 90 &
prevenzione; emorroidi &
  cancer prevention / familiarity; haemorrhoids &
  haemorrhoidal disease &
  \multicolumn{1}{r}{962} &
  \multicolumn{1}{r}{4.647} \\ 91 &
prolasso; emorroidario &
  prolapse; haemorrhoids &
  haemorrhoidal disease &
  \multicolumn{1}{r}{182} &
  \multicolumn{1}{r}{473} \\ 92 &
lynch; sindrome; colon &
  lynch; syndrome; colon &
  lynch syndrome &
  \multicolumn{1}{r}{140} &
  \multicolumn{1}{r}{787} \\ 93 &
addominalgia; proctorragia &
  abdominalgia; proctorragia &
  bleeding &
  \multicolumn{1}{r}{99} &
  \multicolumn{1}{r}{217} \\ 94 &
alvo; ematochezia &
  bowel; haematochezia &
  bleeding &
  \multicolumn{1}{r}{116} &
  \multicolumn{1}{r}{251} \\ 95 &
alvo; sangue; occulto; fecale &
  bowel; blood; occult; faecal &
  bleeding &
  \multicolumn{1}{r}{175} &
  \multicolumn{1}{r}{382} \\ 96 &
alvo; sanguinamento; irregolare; digestivo &
  bowel; bleeding; irregular; digestive &
  bleeding &
  \multicolumn{1}{r}{703} &
  \multicolumn{1}{r}{4.915} \\ 97 &
ematiche; perdite &
  bleeding; leakage &
  bleeding &
  \multicolumn{1}{r}{151} &
  \multicolumn{1}{r}{639} \\ 98 &
ematochezia; positivo; sanguinamento; natura; determinare &
  haematochezia; positive; bleeding; nature; determine &
  bleeding &
  \multicolumn{1}{r}{760} &
  \multicolumn{1}{r}{12.945} \\ 99 &
ematochezia; addominalgia; addominalgie &
  haematochezia; abdominalgia; abdominalgia &
  bleeding &
  \multicolumn{1}{r}{61} &
  \multicolumn{1}{r}{134} \\ 100 &
ematochezia; ematochezie &
  haematochezia; haematochezia &
  bleeding &
  \multicolumn{1}{r}{500} &
  \multicolumn{1}{r}{5.307} \\ 101 &
melena; riferita; episodio &
  melaena; reported; episode &
  bleeding &
  \multicolumn{1}{r}{182} &
  \multicolumn{1}{r}{953} \\ 102 &
nuovi; anticoagulanti; orali; terapia; proctorragia; anemia; paziente &
  new; anticoagulants; oral; therapy; proctorragia; anaemia; patient &
  bleeding &
  \multicolumn{1}{r}{66} &
  \multicolumn{1}{r}{152} \\ 103 &
occulto; alvo; sangue; positivo &
  occult; alvo; blood; positive &
  bleeding &
  \multicolumn{1}{r}{41} &
  \multicolumn{1}{r}{86} \\ 104 &
occulto; fecale; sangue; anemia &
  occult; faecal; blood; anaemia &
  bleeding &
  \multicolumn{1}{r}{824} &
  \multicolumn{1}{r}{1.917} \\ 105 &
occulto; sangue &
  occult; blood &
  bleeding &
  \multicolumn{1}{r}{4.459} &
  \multicolumn{1}{r}{12.755} \\ 106 &
occulto; sangue; positivo &
  occult; blood; positive &
  bleeding &
  \multicolumn{1}{r}{33} &
  \multicolumn{1}{r}{75} \\ 107 &
oncologica; prevenzione &
  oncological; cancer prevention / familiarity &
  bleeding &
  \multicolumn{1}{r}{2.346} &
  \multicolumn{1}{r}{4.752} \\ 108 &
proctorragia &
  proctorrhagia &
  bleeding &
  \multicolumn{1}{r}{279} &
  \multicolumn{1}{r}{1.236} \\ 109 &
proctorragia; diverticolosi &
  proctorrhagia; diverticulosis &
  bleeding &
  \multicolumn{1}{r}{152} &
  \multicolumn{1}{r}{328} \\ 110 &
proctorragia; emorroidi; episodi &
  proctorrhagia; haemorrhoids; episodes &
  bleeding &
  \multicolumn{1}{r}{571} &
  \multicolumn{1}{r}{1.439} \\ 111 &
proctorragia; rettorragia &
  proctorragia; rectorrhagia &
  bleeding &
  \multicolumn{1}{r}{4.219} &
  \multicolumn{1}{r}{13.288} \\ 112 &
proctorragie; emorroidi &
  proctorragia; haemorrhoids &
  bleeding &
  \multicolumn{1}{r}{65} &
  \multicolumn{1}{r}{220} \\ 113 &
prolasso; proctorragia; emorroidario &
  prolapse; proctorragia; haemorrhoid &
  bleeding &
  \multicolumn{1}{r}{46} &
  \multicolumn{1}{r}{98} \\ 114 &
rettorragia; colon &
  rectorrhagia; colon &
  bleeding &
  \multicolumn{1}{r}{48} &
  \multicolumn{1}{r}{104} \\ 115 &
rettorragia; episodi &
  rectorrhagia; episodes &
  bleeding &
  \multicolumn{1}{r}{408} &
  \multicolumn{1}{r}{1.167} \\ 116 &
rettorragia; tumore; colon &
  rectorrhagia; cancer; colon &
  bleeding &
  \multicolumn{1}{r}{74} &
  \multicolumn{1}{r}{157} \\ 117 &
sangue; episodi; presenza &
  blood; episodes; presence &
  bleeding &
  \multicolumn{1}{r}{330} &
  \multicolumn{1}{r}{1.647} \\ 118 &
sangue; occulto &
  blood; occult &
  bleeding &
  \multicolumn{1}{r}{263} &
  \multicolumn{1}{r}{619} \\ 119 &
sangue; occulto; anemizzazione; positivo &
  blood; occult; anaemia; positive &
  bleeding &
  \multicolumn{1}{r}{122} &
  \multicolumn{1}{r}{292} \\ 120 &
sangue; occulto; calprotectina; fecale; positivo &
  blood; occult; calprotectin; faecal; positive &
  bleeding &
  \multicolumn{1}{r}{25} &
  \multicolumn{1}{r}{64} \\ 121 &
sangue; occulto; fecale &
  blood; occult; faecal &
  bleeding &
  \multicolumn{1}{r}{2.525} &
  \multicolumn{1}{r}{7.258} \\ 122 &
sangue; occulto; fecale; anemizzazione &
  blood; occult; faecal; anaemia &
  bleeding &
  \multicolumn{1}{r}{224} &
  \multicolumn{1}{r}{494} \\ 123 &
sangue; occulto; fecale; calo; ponderale &
  blood; fecal; occult; weight-loss &
  bleeding &
  \multicolumn{1}{r}{140} &
  \multicolumn{1}{r}{312} \\ 124 &
sangue; occulto; fecale; positivita &
  blood; fecal; occult; positive &
  bleeding &
  \multicolumn{1}{r}{198} &
  \multicolumn{1}{r}{529} \\ 125 &
sangue; occulto; fecale; positivo; terapia; anemia &
  blood; faecal; occult; positive; therapy; anaemia &
  bleeding &
  \multicolumn{1}{r}{38} &
  \multicolumn{1}{r}{96} \\ 126 &
sangue; occulto; fecale; tumore &
  blood; occult; faecal; cancer &
  bleeding &
  \multicolumn{1}{r}{171} &
  \multicolumn{1}{r}{364} \\ 127 &
sangue; occulto; polipectomia &
  blood; occult; polypectomy &
  bleeding &
  \multicolumn{1}{r}{105} &
  \multicolumn{1}{r}{250} \\ 128 &
sangue; occulto; positivo &
  blood; occult; positive &
  bleeding &
  \multicolumn{1}{r}{69} &
  \multicolumn{1}{r}{151} \\ 129 &
sangue; stipsi; occulto; fecale &
  blood; constipation; occult; faecal &
  bleeding &
  \multicolumn{1}{r}{163} &
  \multicolumn{1}{r}{357} \\ 130 &
stipsi; ematochezia &
  constipation; haematochezia &
  bleeding &
  \multicolumn{1}{r}{91} &
  \multicolumn{1}{r}{190} \\ 131 &
stipsi; proctorragia &
  constipation; proctorragia &
  bleeding &
  \multicolumn{1}{r}{156} &
  \multicolumn{1}{r}{347} \\ 132 &
adenomatosi; polipi &
  adenomatous; polyps &
  polyps &
  \multicolumn{1}{r}{49} &
  \multicolumn{1}{r}{116} \\ 133 &
asportato; colon; polipo &
  removed; colon; polyp &
  polyps &
  \multicolumn{1}{r}{182} &
  \multicolumn{1}{r}{424} \\ 134 &
asportazione; polipo &
  removal; polyp &
  polyps &
  \multicolumn{1}{r}{457} &
  \multicolumn{1}{r}{1.052} \\ 135 &
colon; polipi &
  colon; polyps &
  polyps &
  \multicolumn{1}{r}{2.079} &
  \multicolumn{1}{r}{6.733} \\ 136 &
exeresi; polipi; pregressa &
  exeresis; polyps; previous &
  polyps &
  \multicolumn{1}{r}{71} &
  \multicolumn{1}{r}{174} \\ 137 &
follow; up; polipectomia &
  follow; up; polypectomy &
  polyps &
  \multicolumn{1}{r}{133} &
  \multicolumn{1}{r}{333} \\ 138 &
polipectomia; colon &
  polypectomy; colon &
  polyps &
  \multicolumn{1}{r}{89} &
  \multicolumn{1}{r}{236} \\ 139 &
polipectomia; controllo &
  polypectomy; control &
  polyps &
  \multicolumn{1}{r}{516} &
  \multicolumn{1}{r}{2.263} \\ 140 &
polipectomia; endoscopica; pregressa &
  polypectomy; endoscopic; previous &
  polyps &
  \multicolumn{1}{r}{109} &
  \multicolumn{1}{r}{356} \\ 141 &
polipectomia; pregressa; addominalgia &
  polypectomy; previous; abdominalgia &
  polyps &
  \multicolumn{1}{r}{49} &
  \multicolumn{1}{r}{111} \\ 142 &
polipectomia; sigma &
  polypectomy; sigma &
  polyps &
  \multicolumn{1}{r}{80} &
  \multicolumn{1}{r}{177} \\ 143 &
polipectomia; tumore; colon &
  polypectomy; cancer; colon &
  polyps &
  \multicolumn{1}{r}{32} &
  \multicolumn{1}{r}{65} \\ 144 &
polipi; adenomi; sorveglianza; pazienti; serrati &
  polyps; adenomas; surveillance; patients; serratus &
  polyps &
  \multicolumn{1}{r}{932} &
  \multicolumn{1}{r}{23.390} \\ 145 &
polipi; colon &
  polyps; colon &
  polyps &
  \multicolumn{1}{r}{610} &
  \multicolumn{1}{r}{15.287} \\ 146 &
polipi; controllo &
  polyps; control &
  polyps &
  \multicolumn{1}{r}{365} &
  \multicolumn{1}{r}{4.950} \\ 147 &
polipi; intestinali &
  polyps; intestinal &
  polyps &
  \multicolumn{1}{r}{186} &
  \multicolumn{1}{r}{881} \\ 148 &
polipi; pregressi &
  polyps; previous &
  polyps &
  \multicolumn{1}{r}{333} &
  \multicolumn{1}{r}{1.128} \\ 149 &
polipo; adenomatoso &
  polyps; adenomatous &
  polyps &
  \multicolumn{1}{r}{44} &
  \multicolumn{1}{r}{104} \\ 150 &
polipo; anale &
  polyp; anal &
  polyps &
  \multicolumn{1}{r}{123} &
  \multicolumn{1}{r}{340} \\ 151 &
polipo; colon &
  polyp; colon &
  polyps &
  \multicolumn{1}{r}{168} &
  \multicolumn{1}{r}{5.634} \\ 152 &
poliposi; colon &
  polyposis; colon &
  polyps &
  \multicolumn{1}{r}{711} &
  \multicolumn{1}{r}{1.847} \\ 153 &
poliposi; intestinale &
  polyposis; intestinal &
  polyps &
  \multicolumn{1}{r}{231} &
  \multicolumn{1}{r}{1.876} \\ 154 &
poliposi; pregressa &
  polyposis; previous &
  polyps &
  \multicolumn{1}{r}{53} &
  \multicolumn{1}{r}{119} \\ 155 &
poliposi; sospetta &
  polyposis; suspected &
  polyps &
  \multicolumn{1}{r}{263} &
  \multicolumn{1}{r}{875} \\ 156 &
pregressa; polipectomia &
  previous; polypectomy &
  polyps &
  \multicolumn{1}{r}{981} &
  \multicolumn{1}{r}{6.133} \\ 157 &
pregressa; poliposi &
  previous; polyposis &
  polyps &
  \multicolumn{1}{r}{412} &
  \multicolumn{1}{r}{2.208} \\ 158 &
pregressa; sigma; asportazione; polipo &
  previous; sigma; removal; polyp &
  polyps &
  \multicolumn{1}{r}{79} &
  \multicolumn{1}{r}{165} \\ 159 &
pregresso; polipo &
  previous; polyp &
  polyps &
  \multicolumn{1}{r}{322} &
  \multicolumn{1}{r}{3.204} \\ 160 &
proctorragia; poliposi; colon &
  proctorragia; polyposis; colon &
  polyps &
  \multicolumn{1}{r}{98} &
  \multicolumn{1}{r}{228} \\ 161 &
sigma; polipo &
  sigma; polyp &
  polyps &
  \multicolumn{1}{r}{272} &
  \multicolumn{1}{r}{621} \\ 162 &
stipsi; ostinata; poliposi &
  constipation; obstinate; polyposis &
  polyps &
  \multicolumn{1}{r}{38} &
  \multicolumn{1}{r}{81} \\ 163 &
alvo; familiarita; tumore &
  alvo; familial; cancer &
  cancer prevention / familiarity &
  \multicolumn{1}{r}{212} &
  \multicolumn{1}{r}{604} \\ 164 &
anamnesi; tumore; colon; familiare &
  anamnesis; cancer; colon; familial &
  cancer prevention / familiarity &
  \multicolumn{1}{r}{200} &
  \multicolumn{1}{r}{3.358} \\ 165 &
colon; familiarita; tumore &
  colon; familial; cancer &
  cancer prevention / familiarity &
  \multicolumn{1}{r}{302} &
  \multicolumn{1}{r}{6.530} \\ 166 &
diagnosi; precoce; prestaz; diagnost; tumori &
  early; diagnosis; performance; diagnostics; cancers &
  cancer prevention / familiarity &
  \multicolumn{1}{r}{106} &
  \multicolumn{1}{r}{1.378} \\ 167 &
familiarita; tumore; colon &
  familial; cancer; colon &
  cancer prevention / familiarity &
  \multicolumn{1}{r}{1.137} &
  \multicolumn{1}{r}{3.286} \\ 168 &
famigliarita; tumore; colon &
  familial; cancer; colon &
  cancer prevention / familiarity &
  \multicolumn{1}{r}{151} &
  \multicolumn{1}{r}{388} \\ 169 &
familiarita; colon; neoplasia &
  familiarity; colon; neoplasm &
  cancer prevention / familiarity &
  \multicolumn{1}{r}{1.160} &
  \multicolumn{1}{r}{3.450} \\ 170 &
familiarita; prevenzione &
  familiarity; cancer prevention / familiarity &
  cancer prevention / familiarity &
  \multicolumn{1}{r}{306} &
  \multicolumn{1}{r}{1.114} \\ 171 &
oncologica; familiarita; prevenzione; tumore &
  oncological; familial; cancer prevention / familiarity; cancer &
  cancer prevention / familiarity &
  \multicolumn{1}{r}{195} &
  \multicolumn{1}{r}{453} \\ 172 &
prestaz; diagnost; diagnosi; precoce; tumori &
  prior; diagnosis; early; diagnosis; cancers &
  cancer prevention / familiarity &
  \multicolumn{1}{r}{55} &
  \multicolumn{1}{r}{136} \\ 173 &
prestaz; diagnost; diagnosi; precoce; tumori; tumore; familiarita; colon &
  service; diagnosis; early; diagnosis; cancers; cancer; familial; colon &
  cancer prevention / familiarity &
  \multicolumn{1}{r}{192} &
  \multicolumn{1}{r}{418} \\ 174 &
prestaz; diagnost; diagnosi; sangue; occulto; precoce; tumori &
  performance; diagnosis; diagnosis; occult; blood; early; cancers &
  cancer prevention / familiarity &
  \multicolumn{1}{r}{49} &
  \multicolumn{1}{r}{126} \\ 175 &
tumore; anamnesi; familiare &
  cancer; history; family history &
  cancer prevention / familiarity &
  \multicolumn{1}{r}{280} &
  \multicolumn{1}{r}{3.603} \\ 176 &
tumore; colon; familiarita &
  cancer; colon; familial &
  cancer prevention / familiarity &
  \multicolumn{1}{r}{618} &
  \multicolumn{1}{r}{4.379} \\ 177 &
tumore; colon; familiarita' &
  cancer; colon; familial &
  cancer prevention / familiarity &
  \multicolumn{1}{r}{401} &
  \multicolumn{1}{r}{5.010} \\ 178 &
tumore; colon; proctorragia; familiarita &
  cancer; colon; proctorship; familiarity &
  cancer prevention / familiarity &
  \multicolumn{1}{r}{127} &
  \multicolumn{1}{r}{274} \\ 179 &
tumore; familiarita &
  cancer; familiarity &
  cancer prevention / familiarity &
  \multicolumn{1}{r}{3.709} &
  \multicolumn{1}{r}{8.696} \\ 180 &
tumore; intestino; familiarita &
  cancer; intestine; familiarity &
  cancer prevention / familiarity &
  \multicolumn{1}{r}{64} &
  \multicolumn{1}{r}{412} \\ 181 &
asportazione; adenoma; esiti &
  removal; adenoma; outcome &
  surgery &
  \multicolumn{1}{r}{45} &
  \multicolumn{1}{r}{91} \\ 182 &
esiti; resezione; retto &
  outcomes; resection; rectum &
  surgery &
  \multicolumn{1}{r}{57} &
  \multicolumn{1}{r}{47} \\ 183 &
pregressa; asportazione &
  previous; removal &
  surgery &
  \multicolumn{1}{r}{921} &
  \multicolumn{1}{r}{105} \\ 184 &
pregressa; asportazione; adenoma &
  previous; removal; adenoma &
  surgery &
  \multicolumn{1}{r}{180} &
  \multicolumn{1}{r}{293} \\ 185 &
pregressa; emicolectomia &
  previous; hemicolectomy &
  surgery &
  \multicolumn{1}{r}{75} &
  \multicolumn{1}{r}{85} \\ 186 &
renale; trapianto; lista; insufficienza; cronica; dialisi &
  renal; transplantation; list; failure; chronic; dialysis &
  surgery &
  \multicolumn{1}{r}{25} &
  \multicolumn{1}{r}{64} \\ 187 &
trapianto; fegato &
  transplant; liver &
  surgery &
  \multicolumn{1}{r}{86} &
  \multicolumn{1}{r}{112} \\ 188 &
colica; sospetta; lesione; riscontrata; altra; tecnica; indagine &
  colic; suspected; lesion; found; other; technique; investigation &
  suspected colic lesion &
  \multicolumn{1}{r}{179} &
  \multicolumn{1}{r}{3.671} \\ 189 & 
  calo; ponderale &
  fall; weight &
  other &
  \multicolumn{1}{r}{333} &
  \multicolumn{1}{r}{1.022} \\ 190 &
colon; irritabile &
  colon; irritable &
  other &
  \multicolumn{1}{r}{263} &
  \multicolumn{1}{r}{1.432} \\ 191 &
displasia; adenoma; tubulare; grado &
  dysplasia; adenoma; tubular; grade &
  other &
  \multicolumn{1}{r}{148} &
  \multicolumn{1}{r}{316} \\ 192 &
fossa; iliaca &
  fossa; iliac &
  other &
  \multicolumn{1}{r}{24} &
  \multicolumn{1}{r}{54} \\ 193 &
grado; displasia &
  grade; dysplasia &
  other &
  \multicolumn{1}{r}{84} &
  \multicolumn{1}{r}{191} \\ 194 &
prolasso; rettale &
  prolapse; rectal &
  other &
  \multicolumn{1}{r}{57} &
  \multicolumn{1}{r}{122} \\ 195 &
sigma; ispessimento &
  sigma; thickening &
  other &
  \multicolumn{1}{r}{43} &
  \multicolumn{1}{r}{87} \\ 196 &
subocclusione; intestinale &
  sub-occlusion; intestinal &
  other &
  \multicolumn{1}{r}{81} &
  \multicolumn{1}{r}{202} \\
  \bottomrule
\end{longtable}




\end{document}